\documentclass[format=acmsmall, review=false, screen=true]{acmart}

\usepackage{subfigure}
\usepackage[ruled]{algorithm2e} 

\SetAlFnt{\small}
\SetAlCapFnt{\small}
\SetAlCapNameFnt{\small}
\SetAlCapHSkip{0pt}
\IncMargin{-\parindent}

\DeclareMathOperator*{\argmin}{arg\,min}
\DeclareMathOperator*{\argmax}{arg\,max}

\setcopyright{acmcopyright}

\acmDOI{0000001.0000001}

\begin{document}
\title[Machine Learning Methods for Data Association in Multi-Object Tracking]{Machine Learning Methods for Data Association in Multi-Object Tracking}  

\author{Patrick Emami}
\affiliation{%
  \institution{University of Florida}
  \streetaddress{432 Newell Dr}
  \city{Gainesville}
  \state{FL}
  \postcode{32611}
  \country{USA}}
\email{pemami@ufl.edu}

\author{Panos M. Pardalos}
\affiliation{%
  \institution{University of Florida}
  \streetaddress{401 Weil Hall}
  \city{Gainesville}
  \state{FL}
  \postcode{32611}
  \country{USA}}
 \email{p.m.pardalos@gmail.com}
 
\author{Lily Elefteriadou}
\affiliation{%
  \institution{University of Florida}
  \streetaddress{512 Weil Hall}
  \city{Gainesville}
  \state{FL}
  \postcode{32611}
  \country{USA}}
 \email{elefter@ce.ufl.edu}
 
\author{Sanjay Ranka}
\affiliation{%
  \institution{University of Florida}
  \streetaddress{432 Newell Dr}
  \city{Gainesville}
  \state{FL}
  \postcode{32611}
  \country{USA}}
 \email{sanjayranka@gmail.com}
 
\begin{abstract}
Data association is a key step within the multi-object tracking pipeline that is notoriously challenging due to its combinatorial nature. A popular and general way to formulate data association is as the NP-hard multidimensional assignment problem (MDAP). Over the last few years, data-driven approaches to assignment have become increasingly prevalent as these techniques have started to mature. We focus this survey solely on learning algorithms for the assignment step of multi-object tracking, and we attempt to unify various methods by highlighting their connections to linear assignment as well as to the MDAP. First, we review probabilistic and end-to-end optimization approaches to data association, followed by methods that learn association affinities from data. We then compare the performance of the methods presented in this survey, and conclude by discussing future research directions.
\end{abstract}



\maketitle

\renewcommand{\shortauthors}{P. Emami et al.}

\section{Introduction}
The assignment problem is a classic combinatorial optimization problem where the goal is to find a weighted matching within a bipartite graph such that the sum of the weights is minimized. Within the field of computer vision, it is often used as a framework for tackling data association in multi-object tracking. In this survey, we set out to reexamine the data association problem through the lens of assignment problems as a means to abstract away details and to create a clear conceptual framework for unifying the many recently proposed learning-based data association algorithms. Visual multi-object tracking is a highly complex topic, so rather than attempt to provide a comprehensive overview, we instead take a closer look at solely the association step. Later, we will suggest surveys that review other aspects of the complete multi-object tracking problem for the interested reader. In this work we argue that studying how machine learning can be used to solve data association is important for the following reasons. First, modern machine learning methods, particularly convolutional neural networks (CNNs), excel at learning discriminative features from raw sensor inputs for computing similarities between objects, which is an integral step for any data-driven matching task. For example, a recent study by Bergman et al. \cite{bergmann2019tracking} showed that a simple CNN bounding box regressor can be exploited to extend object tracks over time and drastically reduce the number of ID switches, putting into question the efficacy of sophisticated data association algorithms. Second, efficient probabilistic tools for approximate inference over highly structured models, such as those that arise in data association, have long been studied and are useful for dealing with noisy sensor measurements. Finally, there are many promising recent works on applying machine learning to directly solve a variety of combinatorial optimization problems \cite{bengio2018machine}, and it is interesting to ask whether assignment problems can be solved in a similar manner. 

Multi-object tracking with one or more sensors plays a significant role in many surveillance and robotics applications. A tracking algorithm provides higher-level systems with the ability to make real-time decisions based on the state of the surrounding environment and is a core part of many scene understanding frameworks. Within intelligent transportation systems, it can be used for increasing pedestrian safety at traffic intersections \cite{meissner2012real}, moving object awareness for self-driving cars \cite{osep2017combined}, and for traffic surveillance \cite{Roy2011,alldieck2016context,yang2017multiple,jodoin2016tracking}. Multi-object tracking also has a myriad of other applications ranging from general security systems to tracking cells in microscopy images \cite{Liang2013}. There are many sensor modalities that can be used for these applications; the most common are video, radar, and LiDAR. As a motivating example, consider a vision system that tracks vehicles and pedestrians at an urban traffic intersection. The real-time tracking data can be used for adaptive traffic signal control to optimize the flow of traffic at that intersection. However, intersections contain numerous challenges for multi-object tracking. Heavy traffic occupying multiple lanes and unpredictable pedestrian motion makes for a cluttered scene with lots of occlusion, false alarms, and missed detections. Variability in the appearance of targets caused by poor lighting and weather conditions is especially problematic for visual tracking. On the other hand, new technologies such as vehicle-to-infrastructure (V2I) communication enables vehicles to transmit information directly to traffic intersections, augmenting the data collected by traffic cameras and other sensors \cite{djahel2015toward}.

\subsection{Data Association in Multi-Object Tracking}

\begin{figure*}[t]
    \centering
	\subfigure[Linear Assignment \label{fig:LAP_1}]{
	\includegraphics[scale=0.32]{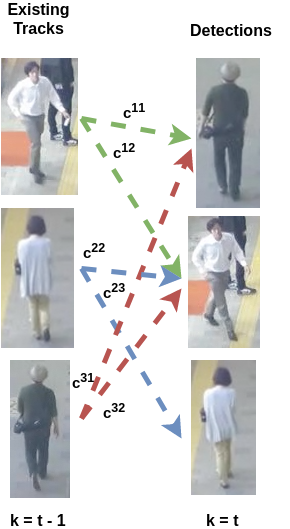}}%
    ~
    \subfigure[Linear Assignment \label{fig:LAP_2}]{
    \includegraphics[scale=0.32]{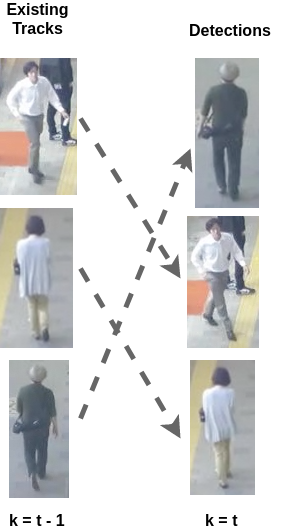}}%
    ~
    \subfigure[Multidimensional Assignment \label{fig:MDAP}]{    \includegraphics[scale=0.32]{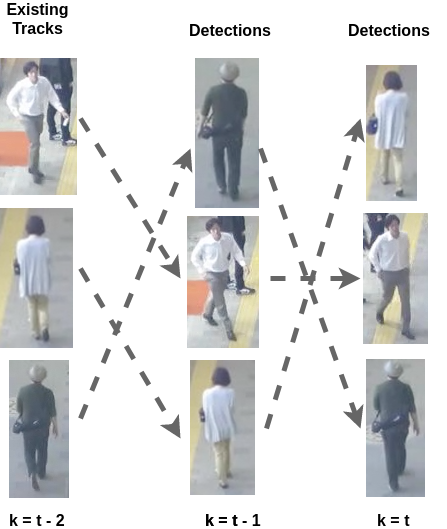}}%
\caption{\textbf{Data association in multi-object tracking}. a) In online tracking, new sensor detections are matched to existing tracks at each time step by solving a linear assignment problem. The assignment hypotheses are the colored, dashed arrows. Each arrow is annotated with the cost $c^{ij}$ of associating track $i$ with detection $j$. b) The optimal linear assignment. Notice how the assignment partitions the set of existing tracks and detections. c) In batch, or offline single-sensor tracking, multiple sets of detections within a sliding window are associated all at once with a set of existing tracks. Here, the sliding window size $T$ is 2 and the optimal assignment is shown. The images are taken from a random video in the MOT Challenge dataset \cite{milan2016mot16}. \label{fig:DA}}
\end{figure*}

\begin{figure*}[t]
\centering
  \subfigure[\label{fig:MS-T2TA-a}]{
  \includegraphics[scale=0.15]{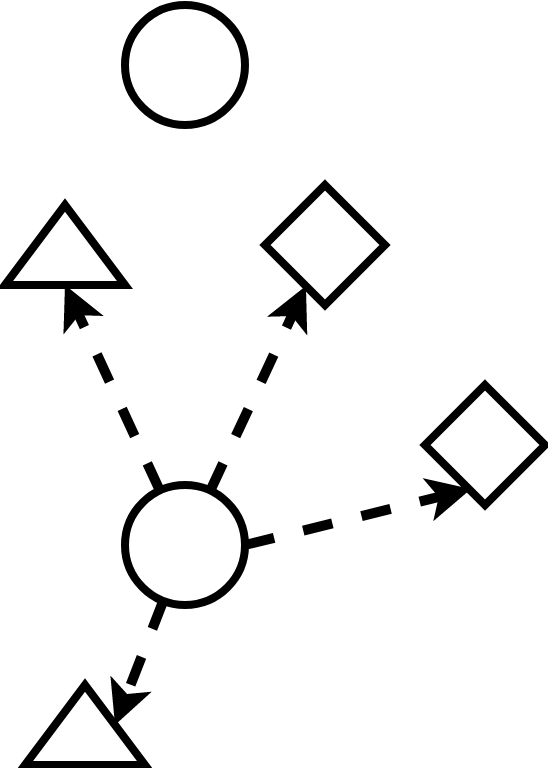}}\hspace{1in}
  \subfigure[\label{fig:MS-T2TA-b}]{
  \includegraphics[scale=0.15]{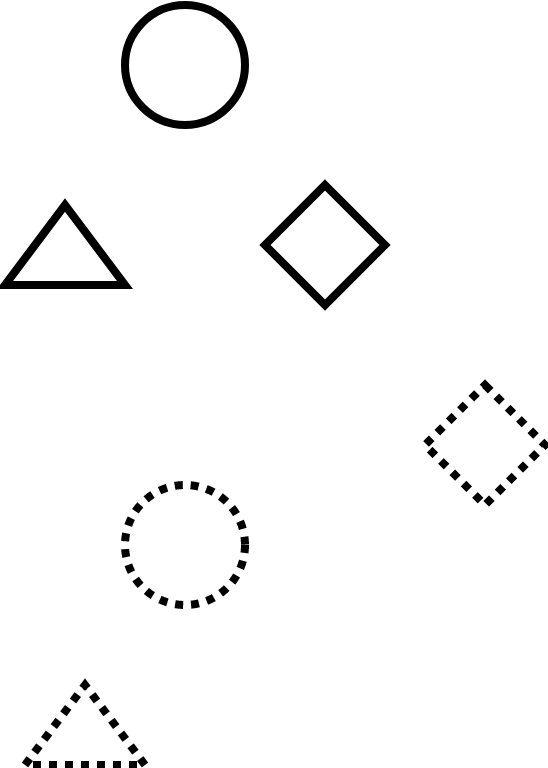}}
\caption{\textbf{Track-to-track Association}. There are three different sensors (circles, triangles, and diamonds) covering the surveillance region, each maintaining two tracks. Suppose there are two ground truth objects. a) The dashed arrows show the possible ways of associating one of the circle tracks with the tracks from the triangle and diamond sensors. b) The best track-to-track association hypothesis. The shapes with solid lines show all tracks, one per sensor, that have been assigned together as having originated from the same ground truth object. Likewise for the shapes with dotted lines. The solution effectively partitions each sensor's track lists.\label{fig:t2ta}}
\end{figure*}

At the core of multi-object tracking lies the measurement-to-track and track-to-track association problems. The goal of measurement-to-track association is to identify a correspondence between a collection of new sensor measurements and preexisting tracks (Figure~\ref{fig:DA}). New measurements can be generated by previously undetected targets, so care must be taken to not erroneously assign one of these measurements to a preexisting track. Likewise, the measurements that stem from clutter within the surveillance region must be identified to avoid false alarms. When there are multiple sensors, there is also the additional problem of track-to-track association. This problem seeks to find a correspondence between tracks that are generated by different sensors (Figure~\ref{fig:t2ta}). Once the optimal assignment of the multi-sensor tracks has been found, all of the tracks assigned to a single track can be combined to produce the final estimate of that track's state. The sensors might be homogeneous or heterogeneous; in the latter case, the problem becomes even harder as the sensors could produce vastly different types of data.
\par 
Broadly speaking, algorithms for solving these two association tasks can be classified as either single-scan, multi-scan, or batch. A single-scan algorithm only uses measurement or track information from the most recent time step, whereas multi-scan algorithms use information from previous and/or future time steps. Batch, or offline multi-object tracking, is an extreme version of multi-scan where the entire sequence is available. Online multi-object tracking operates on one or a few of the most recent scans at a time. Generally, multi-scan methods are preferable in situations where the objects of interest are closely spaced and there are a lot of false alarms and missed detections. However, delaying the association to leverage future information negatively affects the real-time capabilities of the tracker. The accuracy and precision of the tracks produced by multi-scan methods are usually superior, and they offer fewer track ID switches, track breaks, and missed targets \cite{poore2006some}. Naturally, multi-scan methods are more computationally expensive and difficult to implement than their single-scan counterparts. The majority of the algorithms we will discuss in this survey are online algorithms, as offline algorithms typically involve sophisticated global optimization that as of yet is not data-driven.

\par
See Table~\ref{taxonomy of assgn problems} for a categorization of the various data association problems mapped onto assignment problems. The easiest to solve is the bipartite matching or linear assignment problem (LAP), which seeks to match $m$ tracks to $n$ detections. Usually, the problem is constrained so that each track is assigned to exactly one measurement, but measurements are allowed to not be assigned (i.e., false alarms) or to be assigned to a "dummy track" (i.e., a missed detection). For multidimensional data association, e.g., the multi-scan extension of the aforementioned linear assignment problem, extra constraints ensure that each sensor measurement at each time step is assigned to a track exactly once. Unfortunately, the MDAP is NP-hard for dimensions $\geq 3$, whereas there exist many polynomial-time algorithms for the LAP such as the Hungarian method \cite{munkres1957algorithms}.
We will formulate these problems more rigorously in Section~\ref{sec: problem form}. 

\begin{table}[t]
  \caption{\textbf{Taxonomy of assignment problems in multi-object tracking}. LAP~:= linear assignment problem and MDAP~:= multidimensional assignment problem. The algorithms presented in this survey are mostly for solving the various MDAPs encountered in multi-object tracking, and are generally applicable (with modification) to both measurement-to-track and track-to-track association.}
  \label{taxonomy of assgn problems}
  \centering
  \begin{tabular}{lll}
    \toprule
    & \textbf{Measurement-to-Track Association}     & \textbf{Track-to-Track Association} \\
    \midrule
    \textbf{Single-Scan}   & LAP (1-2 sensors), MDAP ($\geq 3$ sensors)  & LAP (2 sensors), MDAP ($\geq 3$ sensors) \\
    \textbf{Multi-Scan}    & MDAP ($\geq 1$ sensors) & MDAP ($\geq 2$ sensors)  \\
    \bottomrule
  \end{tabular}
\end{table}

\subsection{Comparison with Related Surveys}

There are several related surveys to this one and in this section we will highlight their main differences with ours. Both Poore \cite{poore1994multidimensional} and Poore et al. \cite{poore2006some} provide detailed treatments of how assignment problems are useful for multi-object tracking. They only go so far as to frame assignment problems in the context of multi-object tracking. There are a number of excellent general surveys on multi-object tracking \cite{luo2014multiple,yilmaz2006object}; however, their focus is on all aspects of a multi-object tracking solution and they do not have any emphasis on machine learning methods. A survey on appearance matching in camera-based multi-object tracking discusses machine learning methods for improving data association, but it does not cover the recent advances in deep learning that have become ubiquitous in the computer vision tracking community \cite{li2013survey}. The survey by Ciaparrone et al. \cite{ciaparrone2019deep} provides a general overview of deep learning in multi-object tracking.

\subsection{Overview of MOT Benchmarks}

\label{sec: benchmarks}
In this section, we will briefly review the standard multi-object tracking benchmarks. Perhaps the most popular visual-based multi-object tracking set of benchmarks are the MOT challenges. The MOT15 challenge was first released in 2014 and consists of 22 video sequences of pedestrians \cite{leal2015motchallenge}. Since then, the MOT16 and MOT17 challenges have been released, with each release also improving upon the annotation protocol and ground truth quality of the former \cite{milan2016mot16}. These datasets are useful when proposing general improvements to multi-object tracking algorithms since results from many of the state-of-the-art trackers are publicly available for comparison. For an empirical comparison of state-of-the-art trackers on the MOT17 benchmark, see Leal-Taix{\'e} et al. \cite{leal2017tracking}. A more recent comparison that focuses on various deep learning based trackers is available in Ciaparrone et al. \cite{ciaparrone2019deep}.  The MOT datasets are particularly challenging because scenes are filmed from both static and moving vantage points, the density of the crowds of pedestrians is varied, and the appearances of pedestrians drastically changes between sequences. Previously, the PETS \cite{ellis2010pets2010}, TUD Stadtmitte \cite{andriluka2010monocular}, and ETH Pedestrian \cite{eth_biwi_00534} datasets were widely used as benchmarks. These offer a wide variety of multi-view, indoor, and outdoor scenes, and are still useful for training and testing, despite being less frequently used to assess state-of-the-art performance in recent works.

Other datasets of note include the KITTI benchmark \cite{Geiger2012CVPR}, which is is focused on challenges for autonomous driving in urban environments, and contains many tasks beyond multi-object tracking such as odometry, lane estimation, and orientation estimation. 
The UA-DETRAC benchmark \cite{DETRAC:CoRR:WenDCLCQLYL15} is a large-scale traffic surveillance benchmark of 10 hours of video that was recorded at 24 different locations in China, and contains over 8,250 vehicles that were manually annotated. For multi-sensor traffic surveillance, the Ko-PER intersection dataset \cite{strigel2014ko} offers 6 sequences collected with multiple cameras and laser scanners; however, only 2 sequences currently have ground-truth labels.

\subsection{Roadmap}
\begin{figure}
\centering
\includegraphics[scale=0.6]{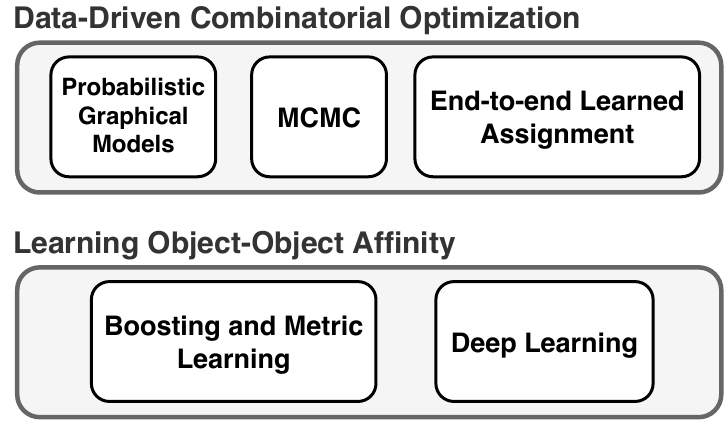}
\caption{Our categorization of machine learning methods for data association. \label{survey tree}}
\end{figure}

Our presentation of data-driven techniques for solving data association is split into two main sections. The first is focused on the combinatorial optimization aspect of the problem, and the second is concerned with learning features for the assignment cost function. Prior to this, in Section~\ref{sec: problem form} we carefully present the connections between data association and assignment problems in multi-object tracking. Section~\ref{sec:opt} will present techniques for finding optimal assignments, with a focus on probabilistic and data-driven algorithms. Then, in Section~\ref{sec: learning_asgn_costs} we present multiple methods for learning features for data association. This presentation is split between algorithms used in multi-object tracking prior to and after the introduction of deep learning. Section~\ref{sec: compare} includes a performance comparison of methods highlighted in this survey, and Section~\ref{sec: conclusions} contains the conclusions. For a visual representation of the organization of the technical contribution of the survey, see Figure~\ref{survey tree}. 

\section{Data Association as Assignment}
\label{sec: problem form}
We will first formally introduce the linear assignment problem (LAP) in the context of single-sensor data association and track-to-track association with two sensors. Following this, we will examine certain MDAP formulations for data association problems.

\subsection{Linear Assignment}

Consider a scenario where there are $m$ existing tracks and $n$ new sensor measurements at time $k$, $k = 1,...,T$. We assume that there is a matrix $C_k \in \mathbb{R}^{m \times n}$, with entries $c^{ij}_k \in C$ representing the cost of assigning measurement $j$ to track $i$ at time $k$ (Figures \ref{fig:LAP_1} and \ref{fig:LAP_2}). The goal is to find the optimal assignment of measurements to tracks so that the total assignment cost is minimized. Using binary decision variables $x^{ij} \in \{0, 1\}$ to represent an assignment of a measurement to a track, we end up with a 0-1 integer program
\begin{equation}
\label{eq:linear-cost}
\min_{x \in X} \sum_{i=1}^m \sum_{j=1}^n c_k^{ij} x^{ij}
\end{equation}
with constraints
\begin{equation}
\begin{aligned}
\sum_{i=1}^{m} x^{ij} &= 1, \hspace{3ex} j = 1, ..., n\\
\sum_{j=1}^{n} x^{ij} &= 1, \hspace{3ex} i = 1, ..., m
\end{aligned}
\end{equation}
where $x \in X$ is a binary assignment matrix. There are $mn$ constraints forcing the rows and columns of $X$ to sum to 1. Note that $C_k$ is not required to be a square matrix. To capture the fact that some sensor measurements will either be false alarms or missed detections, a dummy track is added to the set of existing tracks, so that $C_k$ is now an $(m+1) \times n$ matrix. The entries in the $(m+1)$\textsuperscript{th} row represent the costs of classifying measurements as false alarms. Missed detections are usually handled by forming validation gates around the $m$ tracks (see \cite{blackman1999design}, Section 6.3). These gates can be used to determine, with some degree of confidence, whether any of the new measurements might have originated from a track. The canonical approach is to use elliptical gates, which are typically computed from the covariance estimates provided by a Kalman Filter. In video-based tracking, a similar tactic is to suppress object detections with low confidence values.

Even though there are $\min(m,n)!$ possible assignments, many polynomial-time algorithms exist for finding the globally optimal assignment matrix. Most famous is the $O(n^3)$ Hungarian algorithm \cite{NAV:NAV3800020109,munkres1957algorithms}. Another popular method is the Auction algorithm, introduced by Bertsekas  \cite{bertsekas1992auction}. These algorithms are fast and are easy to integrate into real-time multi-object tracking solutions. However, by only considering the previous time step when assigning measurements or tracks, we are making a Markovian assumption about the information needed to find the optimal assignment. In situations with lots of clutter, false alarms, missed detections, and occlusion, the performance of these algorithms will significantly deteriorate. Indeed, it may be beneficial to instead use a sliding window of previous and/or future track states to construct assignment costs that model the relationship between tracks and new sensor measurements more accurately. As indicated in Table~\ref{taxonomy of assgn problems}, the single-scan track-to-track association problem with two sensors is also a LAP, where $m$ and $n$ represent the sets of tracks maintained by each sensor. Similar methods for handling false alarms and missed detections in data-association can be used for track-to-track association with uneven sensor track lists. If the assignment costs are known, an optimal track assignment can be found in polynomial-time using one of the previously mentioned algorithms.

Instead of abandoning local data association in favor of more expensive global data association approaches, some have proposed heuristics involving solving a cascade of LAPs \cite{Wojke2017simple,al2018multi}. In particular, DeepSORT \cite{Wojke2017simple} has gained in popularity due to its real-time speed and effective use of deep association features to achieve high quality tracking. 

\subsection{Multidimensional Assignment}

Within the single-sensor and multi-sensor tracking paradigms, there are a few different ways to formulate measurement-to-track and track-to-track association as a MDAP (see Table~\ref{taxonomy of assgn problems}). Each formulation seeks to optimize slightly different criteria, but each solution technique is generally applicable to all of them with minor modifications. We suggest further reading on the MDAP for more details \cite{kammerdiner2008,poore1994multidimensional,blackman1999design}.

\subsubsection{Measurement-to-track association}

We begin by considering the MDAP for measurement-to-track association with one sensor given multiple scans. Let the number of scans, or the temporal sliding window size, be given by $T$. Since the objective is to associate new sensor measurements with a set of existing tracks, the resulting MDAP has $T + 1$-dimensions (Figure \ref{fig:MDAP}). When $T \geq 2$, the assignment problem is NP-hard \cite{kammerdiner2008}.
\par
Let the set of noisy measurements at time $k$ be referred to as \textit{scan} $k$ and be represented by $Z_k = \{z_k^{i}\}$, where $i$ is the $i$\textsuperscript{th} measurement of scan $k$, $i = 1, ..., M_k$. $M_k$ is the number of measurements in each scan, i.e., $\lvert Z_k \rvert = M_k$. The main assumption we are making is that each object is responsible for at most one measurement within each scan. We let $Z^T = \{Z_1, ..., Z_T\}$ represent the collection of all measurements in the sliding window of size $T$. 

Let $\Gamma$ be the set of all possible partitions of the set $Z^T$. We seek an optimal partitioning $\gamma^* \in \Gamma$, also called a hypothesis, of $Z^T$ into tracks. Note that a track is just an ordered set of measurements $\{z_1^{i}, z_2^{i}, ..., z_T^{i} \}$; one measurement from each scan at each time step is attributed to each track. Hence, a partition $\gamma$ represents a valid collection of tracks that adhere to the MDAP constraints. Now, we define $\gamma^j$ to be the $j$\textsuperscript{th} track in $\gamma$. Following this, we can define a cost for each track $\gamma^j$ in a partition as $c_{i_1, i_2, ..., i_T}$, where the indices $i_1, i_2, ..., i_T$ indicate which measurements from each scan belong to this particular track. This represents the cost of track $j$ being assigned measurement $i$ from scan 1, measurement $i$ from scan 2, and so on. Crucially, the multidimensional constraints prevent measurements from being assigned to two different tracks and ensure that each measurement is matched to a track. If we use binary variables $\rho_{i_1, i_2, ..., i_T} \in \{0, 1\}$ to indicate if a track is present in a partition, then we can represent the MDAP objective as
\begin{equation}
\label{eq:SSMDAP}
\min_{\gamma \in \Gamma} \sum_{i_1 = 1}^{M_1} \ldots \sum_{i_T = 1}^{M_T} c_{i_1, i_2, ..., i_T}\rho_{i_1, i_2, ..., i_T}  
\end{equation}
with constraints
\begin{equation}
\label{SSMDAP constraints}
\begin{aligned}
\sum_{i_2 = 1}^{M_1} \ldots \sum_{i_T = 1}^{M_T} \rho_{i_1, i_2, ..., i_T} &= 1; \hspace{1cm} i_1 = 1, ..., M_1 \\
\sum_{i_1 = 1}^{M_1} \ldots \sum_{i_T = 1}^{M_T} \rho_{i_1, i_2, ..., i_T} &= 1; \hspace{1cm} i_2 = 1, ..., M_2 \\
&\vdots \hspace{2cm} \vdots\\
\sum_{i_1 = 1}^{M_1} \ldots \sum_{i_{T-1} = 1}^{M_T-1} \rho_{i_1, i_2, ..., i_T} &= 1; \hspace{1cm} i_T = 1, ..., M_T. \\
\end{aligned}
\end{equation}
The solution $\rho$ to this MDAP is the multidimensional extension of the binary assignment matrix. Simply, one may consider $\rho$ as being a multidimensional array with binary entries, such that the sum along each dimension is 1. Similarly to the LAP, we can augment each scan by including a $z_k^0$ dummy measurement in the set of detections at time $k$ to address false alarms. This is useful for identifying track birth and track death as well, but care should be taken when defining the cost for assigning measurements as false alarms or missed detections to avoid high numbers of false positives and false negatives.
\par
It is common to solve for an approximate solution within a fixed-sized sliding window $T$, then shift the sliding window forward in time by $t < T$ so that the new sliding window overlaps with the old region. This allows for tracks to be linked over time, and it provides a compromise between ``offline" tracking, when $T$ is set to the length of an entire sequence of measurements, and "online" tracking, when $T = 1$.
\par

\subsubsection{Track-to-track association}

The other form of the MDAP we are interested in is multi-sensor association with $S \geq 3$ sensors. This scenario is common in centralized tracking systems, where sensors that are distributed around a surveillance region report raw measurements to a central node \cite{sorokin2009mathematical,boginski2011sensors}. When each sensor sends its local tracks to a central node for track association and fusion, a MDAP must be solved. In this case, the dimensionality of the MDAP is equal to $S$, and hence, is NP-hard. Multi-scan track-to-track association with two sensors is also a MDAP, as well as multi-scan multi-sensor measurement-to-track association (Table~\ref{taxonomy of assgn problems}). 
\par
Following \cite{deb1997generalized}, in this scenario there are $S \geq 3$ sensors, each maintaining a set of local tracks and using a sliding window of size $T \geq 1$. We define $X_k^s = \{x_k^{i,s}\}$, $s = 1, ..., S$, to represent the set of track state estimates produced by sensor $s$ at time $k$. We have $i = 1, ..., N_s$, where $N_s$ is the number of tracks being maintained by sensor $s$ and $x_k^{i,s}$ interpreted as the $i$\textsuperscript{th} track of sensor $s$ at scan $k$. Then, for each sensor, we have $X^{T,s} = \{X_1^s, ..., X_T^s\}$, which represents the collection of track state estimates within the sliding window. We seek an optimal partitioning $\gamma^* \in \Gamma$ of $X^{T} = \{X^{T,1}, ..., X^{T,S}\}$ of tracks over all scans and sensors that minimizes the total assignment cost, and we can define a partial assignment hypothesis in a partition $\gamma$ as $\gamma^l = \{ \{x_1^{j,1}, x_1^{j,2}, ..., x_1^{j,N_s}\}, ..., \{ x_T^{j,1}, x_T^{j,2}, ..., x_T^{j,N_s}\}\}$. In words, this states that the $j$\textsuperscript{th} track of sensor 1 from scan 1, $j$\textsuperscript{th} track of sensor 2 from scan 1, and so on, all correspond to the same underlying track $l$ in scan 1. Likewise, this interpretation extends for all subsequent scans. As a quick example, suppose that there are 3 sensors each maintaining 3 tracks, and that $T = 1$. Then a potential hypothesis $\gamma$, or assignment, is $\{ \{x^{1,1}, x^{2,2}, x^{1,3}\}, \{x^{2,1}, x^{1,2}, x^{2,3}\}, \{x^{1,3}, x^{2,3}, x^{3,3}\} \}$. This hypothesis makes the claim that track 1 from sensor 1, track 2 from sensor 2, and track 1 from sensor 3 all were generated by "true" track 1. The assignments for the other two tracks can be identified similarly. Note that the number of "true" targets in the surveillance region must either be known \textit{a priori} or estimated. Considering the simplest case of $T = 1$, we can write the cost for a partial hypothesis as $c_{i_1,i_2, ..., i_{N_s}}$. Increasing $T$ to include more than one scan corresponds to adding extra dimensions to the problem. We can use binary variables as before, $\rho_{i_1, i_2, ..., i_{N_s}} \in \{0, 1\}$, to indicate whether a particular partial hypothesis is present in $\gamma$. The MDAP can then be written as
\begin{equation}
\label{eq:MSMDAP}
\min_{\gamma \in \Gamma} \sum_{i_1 = 1}^{N_1} \ldots \sum_{i_{N_s} = 1}^{N_s} c_{i_1, i_2, ..., i_{N_s}}\rho_{i_1, i_2, ..., i_{N_s}}  
\end{equation}
with constraints
\begin{equation}
\label{MSMDAP constraints}
\begin{aligned}
\sum_{i_2 = 1}^{N_1} \ldots \sum_{i_{N_s} = 1}^{N_s} \rho_{i_1, i_2, ..., i_{N_s}} &= 1; \hspace{1cm} i_1 = 1, ..., N_1 \\
\sum_{i_1 = 1}^{N_1} \ldots \sum_{i_{N_s} = 1}^{N_s} \rho_{i_1, i_2, ..., i_{N_s}} &= 1; \hspace{1cm} i_2 = 1, ..., N_2 \\
&\vdots \hspace{2cm} \vdots\\
\sum_{i_1 = 1}^{N_1} \ldots \sum_{i_{N_{s-1}} = 1}^{N_{s-1}} \rho_{i_1, i_2, ..., i_{N_s}} &= 1; \hspace{1cm} i_{N_s} = 1, ..., N_s. \\
\end{aligned}
\end{equation}
As with the multi-scan data association problem, the solution takes the form of a multidimensional binary array. As before, the number of potential assignment hypotheses in a MDAP can be reduced with gating. Even with gating, solving a MDAP for real-time tracking is infeasible. An analysis on the number of local minima in MDAPs with random costs shows that it increases exponentially in the number of dimensions \cite{grundel2007number}. Notably, the MDAP is closely related to other NP-Hard combinatorial optimization problems, such as Maximum-Weight Independent Set and Set Packing \cite{collins2012multitarget}. In the next subsection, we will show how the costs can be interpreted as probabilities; this will help motivate the use of approximate inference techniques for finding \textit{maximum a posteriori} (MAP) solutions to MDAPs. However, we will begin our discussion of optimization approaches in Section~\ref{sec:opt} with techniques that do not require any assumptions about the nature of the cost function.

\section{Algorithms for Finding Optimal Assignments}
\label{sec:opt}

We begin by briefly reviewing non-probabilistic optimization algorithms for solving the data association problem. These mostly fall into the category of offline data association. Next, our focus will shift to methods with a machine learning flavor. The techniques discussed in this section are quite general, and in most cases can be used for both the measurement-to-track and track-to-track MDAPs with proper modification. The majority of these algorithms are developed for online MOT. We conclude by reviewing recent progress on end-to-end data association, which attempt to replace the combinatorial aspects of the problem with data-driven methods. 

\subsection{Non-probabilistic Algorithms}
\subsubsection{Search Algorithms}

Heuristically searching through the space of valid solutions within a time limit is an attractive way of ensuring both real-time performance and that a good local optima will be discovered. A search procedure for a MDAP takes as input a problem instance in the form of Equation~\ref{eq:SSMDAP} or Equation~\ref{eq:MSMDAP} and constructs a valid solution $\gamma$ by adding each legal partial assignment incrementally. The most well-known method, the Greedy Randomized Adaptive Search Procedure (GRASP), was originally introduced for multi-sensor multi-object tracking \cite{murphey1997greedy}. 

Other greedy search algorithms have been proposed \cite{perea2011greedy,shapero2016adaptive} based on the semi-greedy track selection (SGTS) algorithm \cite{caponi2004polynomial}. SGTS-based algorithms first perform the usual greedy assignment algorithm step of sorting potential tracks by track score, then they generate a list of candidate hypotheses and return the locally optimal result. The main strength of search algorithms appear to be their simplicity and the extent to which they are embarrassingly parallel. 

For a survey of research on GRASP for optimization, see Resende~\cite{Resende2016}.

\subsubsection{Lagrangian Relaxation}

The multidimensional binary constraints \ref{SSMDAP constraints} and \ref{MSMDAP constraints} pose a significant challenge; a standard technique is to relax the constraints so that a polynomial-time algorithm can be used to find an acceptable sub-optimal solution. The existence of $O(n^3)$ algorithms \cite{NAV:NAV3800020109,munkres1957algorithms,bertsekas1992auction} for the LAP suggests that if the constraints can be relaxed, a reasonably good solution to the MDAP should be obtainable within an acceptable amount of time. Indeed, Lagrangian relaxation algorithms for association in multi-object tracking \cite{deb1993multisensor,deb1997generalized} involve iteratively producing increasingly better solutions to the MDAP by successively solving relaxed LAPs and reinforcing the constraints. 

A parallel implementation of this method for the K-best case was developed \cite{popp1998adaptive,popp2001m}, which enables efficient implementations of MHT algorithms. A variation on this approach using dual decomposition has been proposed as well \cite{lau2011multidimensional}.

Lagrangian relaxation has also been used to convert Equation~\ref{eq:SSMDAP} into a global network flow problem \cite{butt2013multi}. The motivation behind this approach is a desire to incorporate higher-order motion smoothness constraints, beyond what is capable when only considering pairwise costs in multi-scan problems. The minimum-cost network flow problem that results from the relaxation can be solved in polynomial-time; updates to the Lagrange multipliers enforcing the constraints are handled by subgradient methods. In the next subsection, we go into more detail on network optimization, one of the leading approaches to solving multi-object tracking association problems.

\begin{figure}[t]
\includegraphics[scale=0.35]{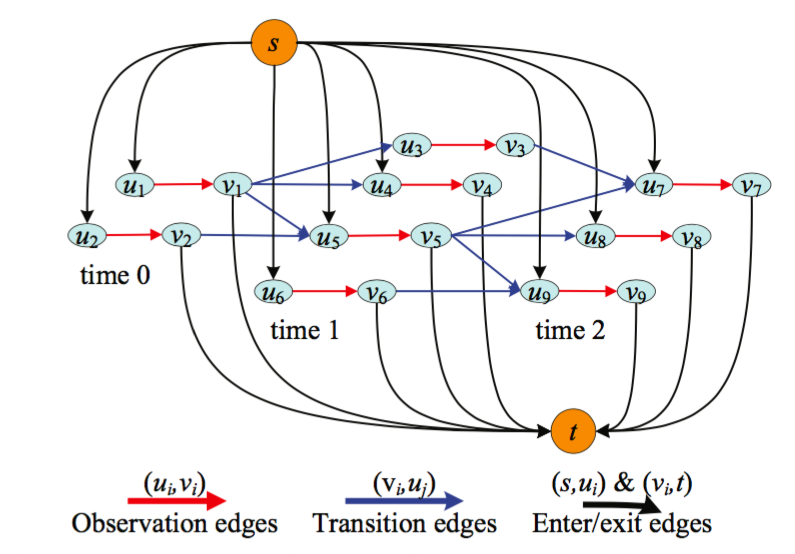}
\caption{A network flow graph for multi-scan data association (three scans depicted). The black arcs represent enter/exit edges for a potential track. The red arcs are measurement/observation edges, and the blue arcs are transition edges between measurements. Reproduced from \cite{zhang2008global} with permission. \label{network-flow}}
\end{figure}

\subsection{Probabilistic Graphical Models}
\subsubsection{Network Optimization}
\label{section:graph-cut}
A popular approach (Equation~\ref{eq:SSMDAP}) in the multi-object tracking computer vision community is to transform the data association problem into finding a minimum-cost network flow \cite{jiang2007linear,zhang2008global,pirsiavash2011globally,berclaz2011multiple,wang2015learning,wang2017tracklet,wu2012coupling,butt2013multi,schulter2017deep,choi2015near,yang2017hybrid}. In the corresponding network, detections at each discrete time step generally become the nodes of the graph, and a complete flow path represents a target track, or trajectory. The amount of flow sent from the source node to the sink node corresponds to the number of targets being tracked, and the total cost of the flow on the network corresponds to the log-likelihood of the association hypothesis. The globally optimal solution to a minimum-cost network flow problem can be found in polynomial-time, e.g., with the push-relabel algorithm. 
\par
Another benefit of using minimum-cost network flow is that the graph can be constructed to significantly reduce the potential number of association hypothesis by limiting transition edges between nodes with a spatiotemporal nearness criteria, similar to gating. Furthermore, occlusion can be explicitly modeled by adding nodes to the graph corresponding to the case where a target is partially or fully occluded by another target for some amount of time. A sliding window approach can be used for real-time performance, rather than using the complete history of previous detections. To help illuminate the mapping from Equation~\ref{eq:SSMDAP} to a network flow problem, we adapt the following equations from Zhang et al. \cite{zhang2008global}, rewritten using the notation from Section~\ref{sec: problem form}.
\par
Recall that we defined a data association hypothesis $\gamma$ as a partitioning of the set of all available measurements $Z^T$. Then, a MAP formulation of the MDAP for data association is given by
\begin{equation}
\label{map network objective}
\begin{aligned}
\gamma^* = & \argmax_{\gamma \in \Gamma} {P(Z^T \mid \gamma)} \prod_{\mathcal{T}_m \in \gamma} P(\mathcal{T}_m) \\
& \textrm{s.t. } {\mathcal{T}_m \cap \mathcal{T}_n = \emptyset, \forall m \neq n}
\end{aligned}
\end{equation}
where the product over tracks in the objective reflects an assumption of track motion independence, and the potentially prohibitive constraint guarantees that no two tracks ever intersect. It is possible to derive the measurement likelihood using Equation~\ref{measurement likelihood}; in Zhang et al. \cite{zhang2008global}, it is factored as $P(Z^T \mid \gamma) = \prod_z P(\{z \in Z^T\} \mid \gamma) $, where each term in this product is a Bernoulli distribution with parameter $\beta$ encoding the probability of false alarm and missed detection. The track probabilities $P(\mathcal{T}_m)$ are modeled as Markov chains to capture track initialization, termination, and state transition probabilities. A network flow graph can now be defined as a graph with source $s$ and sink $t$ as follows. For every measurement $z_k^{i} \in Z^T$, create two nodes $u_r, v_r$, create an arc $(u_r, v_r)$ with cost $c(u_r, v_r)$ and flow $f(u_r, v_r)$, an arc $(s, u_r)$ with cost $c(s, u_r)$ and flow $f(s, u_r)$, and an arc $(v_r, t)$ with cost $c(v_r, t)$ and flow $f(v_r, t)$. For every transition $P(z_{k+1}^{i} \mid z_k^{i}) \neq 0$, create an arc $(v_r, u_s)$ with cost $c(v_r, u_s)$ and flow $f(v_r, u_s)$. An example of such a graph is given in Figure~\ref{network-flow}. The flows $f$ are indicator functions defined by
\begin{equation}
\begin{aligned}
f(s,u_r)  &= \begin{cases} 
   1 & \text{if } \exists \mathcal{T}_m \in \mathcal{T}, \mathcal{T}_m \text{ starts from } u_r \\
   0       & \text{otherwise} \\
  \end{cases} \\
f(v_r, t) &= \begin{cases} 
   1 & \text{if } \exists \mathcal{T}_m \in \mathcal{T}, \mathcal{T}_m \text{ ends at } v_r \\
   0       & \text{otherwise} \\
  \end{cases} \\
f(u_r, v_r)  &= \begin{cases} 
   1 & \text{if } \exists \mathcal{T}_m \in \mathcal{T}, z_k^{i} \in \mathcal{T}_m \\
   0       & \text{otherwise} \\
  \end{cases} \\
f(v_r, u_s) &= \begin{cases} 
   1 & \text{if } \exists \mathcal{T}_m \in \mathcal{T}, z_{k+1}^{i} \text{ comes after } z_k^{i} \text{ in } \mathcal{T}_m \\
   0       & \text{otherwise} \\
  \end{cases}
\end{aligned}
\end{equation}
and the costs are defined as 
\begin{equation}
\label{eq:link-costs}
\begin{aligned}
&c(s, u_r) = -\log P_{\text{start}}(z_k^{i}) & &c(v_r, t) = -\log P_{\text{end}}(z_k^{i}) \\
&c(u_r, v_r) = \log \frac{\beta_r}{1 - \beta_r} & &c(v_r, u_s) = -\log P_{\text{link}}(z_{k+1}^{i} \mid z_k^{i}),
\end{aligned}
\end{equation}
and can be derived by taking the logarithm of Equation~\ref{map network objective}; see Section 3.2 of \cite{zhang2008global} for more details. The minimum cost flow through the network corresponds to the assignment $\gamma^*$ with the maximum log-likelihood.
\par
Quite a few variations on this model have been proposed in the literature.  In one case, a subgraph is created for each track in the surveillance region and occlusion is modeled by adding special nodes to the graphs \cite{jiang2007linear}. A linear programming relaxation with a sliding-window heuristic then enables approximate global solutions to be found in real-time. A limitation of this approach is the requirement of knowing \textit{a priori} the number of tracks in the surveillance region, as well as the poor worst-case complexity of the simplex method. Another work further optimizes the approach introduced in Zhang et al. \cite{zhang2008global} to reduce the run-time complexity \cite{pirsiavash2011globally}. In a more drastic departure from previous works in this direction, the problem has also been formulated as a K-shortest paths through a flow graph \cite{berclaz2011multiple}. One argument against the previously discussed network flow models is that they exhibit an over-reliance on appearance modeling and pairwise costs \cite{collins2012multitarget}. They offer a variation on the network flow approach that uses a more general cost function. In Section~\ref{sec: learning_asgn_costs}, we will go over the details of works that propose a variety of machine learning techniques to obtain the link costs (Equation~\ref{eq:link-costs}) in network flow graphs. Network optimization techniques offer a good trade-off between complexity, ease of implementation, and performance.

\subsubsection{Conditional Random Fields}
\label{section:crf}
Probabilistic graphical models provide us with a powerful set of tools for modeling spatiotemporal relationships amongst sensor measurements in data association and amongst tracks in track-to-track association. Indeed, conditional random fields (CRFs), a class of Markov random fields \cite{Lafferty:2001:CRF:645530.655813}, have been used extensively for solving MDAPs in visual tracking \cite{milan2016multi,yang2012online,yang2011learning,le2016long,choi2015near,osep2017combined}. A CRF is an undirected graphical model, often used for structured prediction tasks, that can represent a conditional probability distribution between sets of random variables. CRFs are well-known for their ability to exploit grid-like structure in the underlying probabilistic model. 
\par
We define a CRF over a graph $G = (V, E)$ with nodes $x_{v \in V} \in X$ such that each node emits a label $y \in Y$. For simplicity of notation, we refer to nodes as $x$ and omit the subscript. The labels take on values from a discrete set, e.g., $\{0, 1\}$; in the context of multi-object tracking, a realization of labels $\mathbf{y}$ usually corresponds to an assignment hypothesis. A key theorem concerning random fields states that the probability distribution being modeled can be written in terms of the cliques $c$ of the graph \cite{hammersley1971markov}. For example, in chain-structured graphs, each pair of nodes and corresponding edge is a clique.
\par
CRFs, like the probabilistic network flow models discussed in the previous subsection, are essentially a tool for modeling probabilistic relationships between a collection of random variables. They require a separate optimization process for handling training and inference (such as the graph cut algorithm \cite{boykov2004experimental} or message passing algorithms). We will focus on presenting how the data association problem is mapped onto a CRF and direct the reader to other sources \cite{boykov2004experimental} for details on how exactly approximate inference is carried out for these models. One of the benefits of using graphical models is that we have the flexibility to construct our graph using either sensor measurements, tracklets (measurements that are partially associated to form a "sub"-track), or full tracks. Tracklets are a common choice for CRFs since they give an attractive hierarchical quality to the tracking solution; low-level measurements are first associated into tracklets via, e.g., the Hungarian algorithm, and then stitched together into full tracks via a CRF. By working at a higher level of abstraction, the original MDAP constraints \ref{SSMDAP constraints} and \ref{MSMDAP constraints} are modified slightly; all that is needed at the higher level is to ensure that each tracklet is only associated to one and only one track. This can also help reduce processing time for running in real-time. 
\par
Each clique $c$ in the graph has a clique potential $\psi_c$ associated with it; usually, the clique potentials are written as the product of unary terms $\psi_s$ and pairwise terms $\psi_{s,t}$. It is common to assume a log-linear representation for the potentials, i.e., $\psi_c = \exp (w_c^\intercal \phi (x, y_c))$. Note that the implied normalization term in Equation~\ref{crf} can be omitted when solving for the maximum-likelihood labeling $\mathbf{y}$ for a particular set of observations $\mathbf{x}$, such that 
\begin{equation}
\begin{aligned}
\label{crf}
P(\mathbf{y} \mid \mathbf{x}, w) &\propto \prod_c \psi_c(y_c \mid \mathbf{x}, w) \\
	& \propto \prod_{s \in V} \psi_s (y_s \mid \mathbf{x}, w) \prod_{s,t \in E} \psi_{s,t} (y_s, y_t \mid \mathbf{x}, w).
\end{aligned}
\end{equation}
Features $\phi$ must be provided (or can be extracted from data with supervised or unsupervised learning) and weights $w$ are learned from data.  The observations $\mathbf{x}$ can be either sensor measurements (for data association) or sensor-level tracks (for track-to-track association). The Markov property of CRFs can be interpreted in the context of multi-object tracking as assuming that the assignment of the observations to tracks within a particular spatiotemporal section of the surveillance region is independent of how they are assigned to tracks elsewhere---conditional on all observations. This adds an aspect of local optimality and, in a way, embeds similar assumptions as a gating heuristic. A solution to Equation~\ref{crf}, i.e., the maximum-likelihood set of labels $\mathbf{y}$, can be used as a solution to the corresponding MDAP. 
\par
As is common with CRFs, the problem of solving for the most likely assignment hypothesis is cast as energy minimization. The objective to minimize is an energy function, computed by summing over the clique potentials; each potential is interpreted as contributing to the energy of the assignment hypothesis. Each clique consists of a set of vertices and edges, where each vertex is a pair of tracklets that could potentially be linked together. The corresponding labels for each vertex take values from the set $\{0, 1\}$ and indicates whether a pair of tracklets are to be linked or not. The energy term for each clique is decomposed into the sum of a unary term for the vertices and a pairwise term for the edges. In one instance, the weights $w$ are learned with the RankBoost algorithm \cite{yang2011learning}. Other techniques for learning the parameters of a CRF that maximize the log-likelihood of the training data include iterative scaling algorithms \cite{Lafferty:2001:CRF:645530.655813} and gradient based techniques. In Section~\ref{sec: learning_asgn_costs}, we will examine the problem of learning weights for assignment costs in more detail. The features used to construct these terms include appearance, motion, and occlusion information, among others. CRF and network optimization-based trackers are by nature global optimizers, and must be run with a temporal sliding-window to get near real-time performance. For example, extensions to the generic CRF formulation have been developed that enable it to run in real-time \cite{yang2012online}.
\par
A particular CRF formulation, Near Online Multi-Target Tracking (NOMT) \cite{choi2015near}, also builds its graph of track hypotheses using tracklets. The novelty of this work is in the use of an affinity measure between detections called the Aggregated Local Flow Descriptor, and in the specific form of the unary and pairwise terms in the energy function of the CRF. Inference in the CRF is sped up by first analyzing the structure of the graphical model so that independent subgraphs can be solved in parallel.

Other variations on the approaches above have been seen as well. In one such work, the energy term of a CRF is augmented with a continuous component to jointly solve the discrete data association and continuous trajectory estimation problems \cite{milan2016multi}. Another study embedded a factor graph in the CRF to add more structure and help model pairwise associations explicitly \cite{heili2014exploiting}. Based on the insight that the size of the bounding box is an indicator of object localization accuracy, asymmetric pairwise terms are added to the CRF that take this idea into account for better uncertainty management \cite{zhou2018deep}. 

In the sequel, we will investigate how factor graphs, the belief propagation inference algorithm, and its variants can be used to solve the MDAP. To summarize, applying CRFs to a specific multi-object tracking problem involves defining how the graphical model will be constructed from the sensor data, specifying an objective function, selecting or learning features for the terms within the objective function, training the model to learn the weights, and then performing approximate inference to extract the predicted assignment hypothesis. 

\subsubsection{Belief Propagation}
\label{section:belief-prop}
In this section, we highlight recent work that formulate the association problems as MAP inference and use belief propagation (BP) or one of its variants to obtain a solution. Chen et al. \cite{chen2006data,chena2009efficient} showed the effectiveness of BP at finding the MAP assignment hypothesis for the single and multi-sensor data association problems. BP is a general message-passing algorithm that can carry out exact inference on tree-structured graphs and approximate inference on graphs with cycles, or "loopy" graphs. The types of graphs under consideration are once again Markov random fields, albeit more general ones than the ones discussed in the previous subsections. Indeed, BP can be used on graphs that model joint distributions $P(\mathbf{x}) = P(x_1, x_2, ..., x_N)$ that can be factorized into a product of clique potentials. As before, the clique potentials are assumed to be factorizable into pairwise terms. Therefore, for cliques $c$, we have
\begin{equation}
\label{bp-1}
\begin{aligned}
P(\mathbf{x}) &\propto \prod_{c} \psi_c(x_c) \\
	& \propto \prod_{s \in V} \psi_s (x_s) \prod_{s,t \in E} \psi_{s,t} (x_s, x_t).
\end{aligned} 
\end{equation}
It is common to use factor graphs to explicitly encode dependencies between variables. A factor graph decomposes a joint distribution into a product of several local functions $f_j(X_j)$, where each $X_j$ is some subset of $\{x_1, x_2, ..., x_N\}$. The graph is bipartite and has nodes $x$ (i.e., discrete random variables) and factors (i.e., dependencies) $f \in \mathcal{F}$, and edges between the nodes and factors. For example, the graph of $g(x_1, x_2, x_3) = f_A(x_1) f_B(x_2, x_3) f_C(x_1, x_3)$ has factors $f_A, f_B$, and $f_C$ and nodes $x_1, x_2, x_3$.
The joint distribution for a factor graph can be written similarly to Equation~\ref{bp-1} as
\begin{equation}
P(\mathbf{x}) \propto \prod_{s \in V} \psi_s (x_s) \prod_{f \in \mathcal{F}} \psi_f (x_{\eta_f}),
\end{equation}
where $\eta_f$ represents the set of nodes $x$ that are connected to factor $f$. 
\par
Parallel message-passing algorithms, such as BP, operate by having each node of the graph iteratively send messages to its neighbors simultaneously. 
We define messages from a node $x_s$ to its neighbors $x_t \in \mathcal{N}(s)$ as $\mu_{s \rightarrow t}(x_s)$. In a factor graph, the set of neighbors $\mathcal{N}(s)$ for a node $x_s$ are its corresponding factors. The max-product algorithm is useful for finding the MAP configuration ${x^*} = \{{x^*}_s \mid s \in V\}$ which corresponds to the best assignment hypothesis $\gamma^*$. In this algorithm, messages are computed recursively in general pairwise Markov random fields by
\begin{equation}
\label{pairwise-LBP}
\mu_{s \rightarrow t} (x_s) = \max_{x_{t}} \bigg \{ \psi(x_{t}) \psi_{s,t} (x_s, x_{t}) \prod_{\xi \in N(t) \text{\textbackslash} s } \mu_{\xi \rightarrow t} (x_{t}) \bigg \}
\end{equation}
and at convergence, each $x^*_s$ can be calculated by 
\begin{equation}
\label{pairwise-LBP-arg}
x^*_s = \argmax_{x_s \in X} \bigg \{ \psi_s(x_s) \prod_{\xi \in \text{ nbr}(s)} \mu_{\xi \rightarrow s} (x_s) \bigg \}
\end{equation}
for neighborhood set $\text{nbr}(s)$. These updates are not guaranteed to converge for graphs with cycles, and even if they do, they may not compute the exact MAP configuration \cite{chen2006data}. See Williams et al. \cite{williams2010convergence} for a proof of convergence of loopy belief propagation (LBP) for data association. LBP simply applies the BP updates repeatedly until the messages all converge; interestingly, LBP has been shown to perform favorably in practice for association tasks \cite{williams2010data,williams2014approximate,meyer2017scalable}. An improvement over the max-product algorithm for LBP is tree-reweighted max-product \cite{wainwright2002map}. This algorithm is used for data association to output a provably optimal MAP configuration or acknowledge failure \cite{chen2006data}. The key idea of the tree-reweighted max-product algorithm is to represent the original problem as a combination of tree-structured problems that share a common optimum \cite{chen2006data}.
\par
To illustrate the use of BP for solving MDAPs, we will present the graphical model formulation from Zhu et al. \cite{zhu2007graphical} for multi-sensor multi-object track-to-track association. The structure of the graphical model is decided on-the-fly by producing sets of independent association clusters consisting of multi-sensor tracks that could plausibly be associated with each other. This is accomplished by computing elliptical gates around each track and clustering together all such tracks whose gates overlap, using e.g., kinematic information. The nodes of the graph are the track state estimates for $T = 1$ and $S \geq 3$ sensors (Section~\ref{sec: problem form}), $\big \{ x^{i,j} \mid x^{i,j} \in X^1 = \{X^{1,1}, X^{1,2}, ..., X^{1,S}\} \big \}$, where each $x^{i,j}$ is the $i$\textsuperscript{th} track state estimate from sensor $j$, $i = 1, ..., N_j$ and $j = 1, ..., S$. Edges only exist between nodes from different sensors when their elliptic gates overlap. A random variable $Y^{i,j}$ corresponding to each node $x^{i,j}$ is defined as a vector of $S-1$ dimensions and stores the indexes of the tracks from the other sensors associated with the $i$\textsuperscript{th} track from sensor $j$. The node potentials are defined as $\psi_{x^{i,j}} (Y^{i,j}) = \exp(\rho)$ where $\rho$ is the sum of pair-wise costs, given by Equation~\ref{sum-pairwise}. Using the notation $Y^{i,j}_k$ to denote the $k$\textsuperscript{th} entry of the $S-1$-dimensional vector $Y^{i,j}$, (the index of the local track from sensor $k$), the edge potentials can be defined to ensure that each track from each sensor is associated once and only once by 
\begin{equation}
\begin{aligned}
\psi_{x^{l,m} \rightarrow x^{n, o}}(Y^{l,m}_n = p, Y^{n,o}_l = q)  &= \begin{cases} 
   0 & p = n, q \neq l \\
   0 & p \neq n, q = l \\
   1 & \text{otherwise}. \\
  \end{cases} \\
\end{aligned}
\end{equation}
If $w^{u,v}$ is the Mahalanobis distance between two tracks $u, v$, then messages between nodes can be initialized as 
\begin{equation}
\begin{aligned}
\mu_{x^{l,m} \rightarrow x^{n,o}} (Y^{n,o}_l = q) &= \begin{cases} 
   \exp(w^{u = (l,m); v = (n,o)}) & \text{ if } q = l \\
   1 & \text{otherwise}. \\
  \end{cases} \\
\end{aligned}
\end{equation}
Then, repeated applications of Equations \ref{pairwise-LBP} and \ref{pairwise-LBP-arg} until the $Y^{i,j}$s converge will produce the MAP solution.
\par
This approach has been extended for an unknown number of targets and multiple sensors \cite{meyer2016tracking} and applied to a multistatic sonar network \cite{meyer2017scalable}. For a general overview of graph techniques for the data association problem, including BP, see Chong \cite{chong2012graph}.

\subsection{Markov Chain Monte Carlo}

A principled approach to sampling from a complex, potentially high-dimensional distribution is Markov Chain Monte Carlo (MCMC). MCMC methods construct a Markov chain on the state space $\mathcal{X}$ whose stationary distribution $\pi^*$ is the target distribution. Decorrelated samples drawn from the chain can be used for approximate inference, i.e., integrating with respect to $\pi^*$. This is useful in the context of assignment problems for multi-object tracking when the goal is to estimate a posterior distribution over assignment hypotheses, from which a MAP hypothesis can be extracted. The Metropolis-Hastings algorithm has been used extensively for data association in single and multi-sensor scenarios \cite{benfold2011stable,pasula1999tracking,oh2004markov,fagot2016improving}. Recently, a Gibbs sampler was derived for efficient implementations of the Labeled Multi-Bernoulli filter, which jointly addresses the data association and state estimation problems for single and multi-sensor scenarios \cite{reuter2017fast,vo2017efficient}. We omit detailed descriptions of the Metropolis-Hastings and Gibbs sampling algorithms, and instead refer the reader to relevant work \cite{vo2017efficient,oh2004markov}.
\par
MCMC is applied to the MDAP for data association (referred to as MCMCDA) and track-to-track association by designating the state space of the Markov chain to be all feasible assignment hypotheses and the stationary distribution of the Markov chain to be the posterior $P(\gamma \mid Z^T)$ or $P(\gamma \mid X^{T})$. A MAP assignment hypothesis $\gamma^*$ for the data association problem is:
\begin{align}
\label{MCMCDA}
P(\gamma \mid Z^T) &\propto P(Z^T \mid \gamma ) \prod_{t=1}^T p_z^{z_t}(1 - p_z)^{c_t} p_d^{d_t} (1 - p_d)^{g_t} \lambda_b^{a_t} \lambda_f^{f_t} \\
\gamma^* &= \argmax_{\gamma} P(\gamma \mid Z^T).
\end{align}
Here, we define the survival probability as $p_z$ and the detection probability as $p_d$. The number of targets at time $t - 1$ is $e_{t-1}$, the number of targets that terminate at time $t$ is $z_t$, and $c_t = e_{t-1} - z_t$ is the number of targets from time $t - 1$ that have not terminated at time $t$. We set $a_t$ as the number of new targets at time $t$, $d_t$ as the number of actual target detections at time $t$, and $g_t = c_t + a_t - d_t$ as the number of undetected targets. Finally, let $f_t = n_t - d_t$ be the number of false alarms, $\lambda_b$ be the birth rate of new objects, and $\lambda_f$ be the false alarm rate. Note that for the general case of unknown numbers of targets, the multi-scan MCMCDA will find an approximate solution of unknown quality at best. A bound on the quality of the approximation for the single-scan fixed target MCMCDA has been derived \cite{oh2004markov}.
\par
A Metropolis-Hastings algorithm for Equation~\ref{MCMCDA} is as follows \cite{oh2004markov}. The proposal distribution $q$ is associated with five types of moves, for a total of eight moves; a birth/death move pair, a split/merge move pair, an extension/reduction move pair, a track update move, and a track switch move. A move is accepted with acceptance probability $A(\gamma, \gamma')$, where 
\begin{equation}
A(\gamma, \gamma') = \min \bigg (1, \frac{\pi(\gamma') q(\gamma', \gamma) }{\pi (\gamma) q(\gamma, \gamma')} \bigg ).
\end{equation}
Assuming a uniform proposal distribution $q$, the proposal distribution terms in the numerator and denominator cancel. The stationary distribution $\pi(\gamma)$ is $P(\gamma \mid Z^T)$ from Equation~\ref{MCMCDA}. Implementation details and descriptions of each type of move can be found in Section V-A of Oh et al. \cite{oh2004markov}. Extensions to this algorithm have been proposed \cite{benfold2011stable} to add a sliding-window version and to reduce the number of types of moves to three. For visual tracking \cite{benfold2011stable}, appearance information is fused with kinematic information to help improve performance. Sparse representations of detections and kinematic information  have been used to define an energy objective that MCMCDA approximately optimizes  \cite{fagot2016improving}. This work deviates from its predecessors by allowing moves to be done not only forward in time, but also backwards to explore the solution space more efficiently. The use of a sliding-window is once again crucial, enabling the trade-off between solution quality and a faster run-time.

\subsection{End-to-end Data Association}
\label{sec:e2e-dl}

\begin{figure}
\includegraphics[scale=0.35]{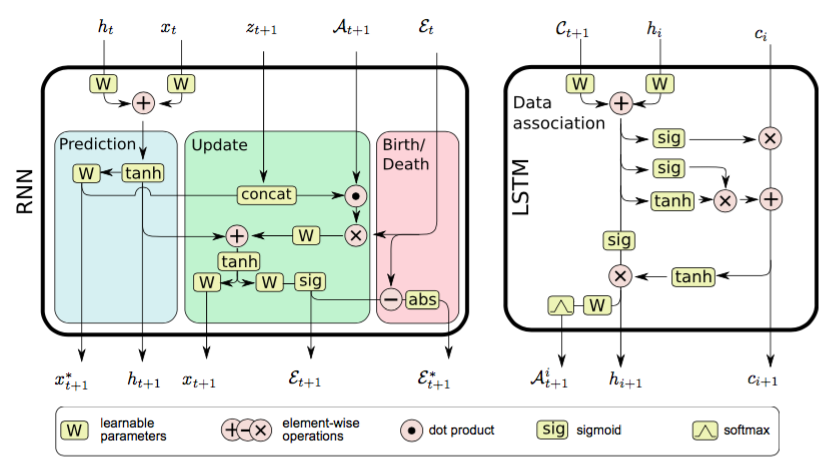}
\caption{An LSTM cell designed for multi-scan single-sensor data association (right). The input at each time step is the matrix of pairwise distances $C_{t+1}$, along with the previous hidden state $h_t$ and cell state $c_t$. The output $A^{i}_{t+1}$ of the data association cell is a vector of assignment probabilities for each target and all available measurements, obtained by a log-softmax operation, and is subsequently fed into the state estimation recurrent network (left). The LSTM's nonlinearities and memory are believed to provide the means for learning efficient solutions to the data association problem. Best viewed in color. Reproduced from \cite{milan2017online} with permission. \label{fig:mtt-with-rnns}}
\end{figure}

Neural networks have a rich history of being used to solve combinatorial optimization problems.
One of the earliest and most influential papers in this line of research, by Hopfield and Tank \cite{hopfield1985neural}, describes how to use Hopfield nets to approximately solve instances of the Traveling Salesperson Problem (TSP).
Despite the controversy associated with their results \cite{smith1999neural}, this work inspired many others to pursue these ideas.
This has lead to the present day, where research on the use of deep neural networks to solve combinatorial optimization problems has started to pick up speed \cite{bengio2018machine}.

Following broad trends within the deep learning research community, many have recently asked whether the data association step in multi-object tracking can be solved in an almost entirely ``end-to-end'' fashion.
In other words, given noisy measurements of the environment, the tracker should directly output filtered tracks, combining the association problem with state estimation into a monolithic learned module.
In this section we will present recent work that attempt to learn the data association step from data using deep learning.

\subsubsection{Data-driven Association}
The Deep Affinity Network \cite{sun2019deep} (DAN) is a deep neural network that explicitly learns the affinity between objects over time. It is trained to predict the optimal linear assignment using ground truth assignment matrices as supervision. Visual features are first extracted from a VGG network then processed by DAN to output a matrix of soft assignments, which finally are stitched into tracks using the Hungarian algorithm. The main insight of this approach is that DAN is able to jointly learn good appearance features as well as features that are highly ``matchable''. They showed equal or better performance on MOT15 and UA-DETRAC with state-of-the-art methods. A closely related tracker is FAMNet \cite{chu2019famnet}, which learns to predict the assignment tensor for the MDAP directly. They use a sliding window to construct a set of hypothesis tracklets, for which an affinity network outputs the affinity tensor for the MDAP. An iterative and differentiable row/column tensor normalization layer is used to directly output the assignment, through which gradients from a loss computed with the ground truth assignment can be backpropagated. Another deep tracker similar to DAN is the Deep Hungarian Network (DHN) \cite{xu2019deepmot}, which also attempts to predict the optimal linear assignment from a cost matrix between measurements and tracks. Interestingly, they derive a differentiable version of the multi-object tracking metrics MOTA and MOTP \cite{bernardin2008evaluating} in order to directly formulate the loss in terms of the MOT metrics given ground truth assignments. The reported performance on the MOT17 benchmark are inferior to DAN, however. 
The Dual Matching Attention network (DMAN) \cite{zhu2018online} augments their data association algorithm by introducing spatial and temporal attention networks which refine candidate assignments. The spatial attention generates dual attention maps to exploit the strengths of discriminative CNN feature embeddings for re-ID, as is commonly done in single-object tracking. 
Tracking by Animation \cite{he2019tracking} is a deterministic unsupervised model that uses attention and memory mechanisms to learn to track using only reconstruction error. It assumes rather simplistic scene compositions to be able to render the predicted scene in a differentiable way. The memory mechanism uses read/write operations to address data association and keep track of which objects have been attended to at each time step. Although their experimental results were mainly on small-scale datasets, this direction is very promising as high-quality labeled data for multi-object tracking is scarce. Finally,  we note that reinforcement learning has been applied successfully to multi-object tracking \cite{xiang2015learning} where a policy is learned over a data association Markov decision process that handles track initialization, maintenance, and removal.

\subsubsection{Recurrent Neural Networks}
An investigation by Ondruska et al. \cite{ondruska2016deeptrack} revealed that a recurrent-convolutional neural network is able to learn to track multiple targets from raw inputs in a synthetic problem without access to labeled training data. Crucially, rather than maximizing the likelihood of the next state of the system at each time step, they modified the cost function to maximize the likelihood at some time $t + n$ in the future to force the network to learn a model of the system dynamics. More recently, they extended this work for use with raw LiDAR data collected by an autonomous vehicle \cite{DequaireIJJ2017}.  Recurrent Autoregressive Networks \cite{fang2017recurrent} was designed as an approach to online multi-object tracking that seeks to incorporate internal and external memory components into a deep learning framework to help handle occlusion and appearance changes. They are able to show that RAN indeed makes use of its external memory to maintain tracks while the targets are occluded. See \cite{sadeghian2017tracking} for a closely related prior work that also explores the use of recurrent neural networks (RNNs). Recently, RNNs were also used to identify track failures (ID switches) within a set of tracklets so as to automatically correct such cases in a post-processing step \cite{ma2018trajectory}. Explicit learning of the assignment problem was attemped by Milan et al.\cite{milan2017online}, where they used deep learning to separately tackle the state estimation and data association problems. They designed a Long Short-Term Memory cell specifically for solving the MDAP in data association (Figure~\ref{fig:mtt-with-rnns}). Despite not using any visual features, their approach achieves reasonable performance relative to other similar systems on the MOT Challenge 2015 dataset \cite{leal2015motchallenge}. 

\subsubsection{Deep Generative Models}
Advances in our ability to train and scale deep generative models such as variational auto-encoders \cite{kingma2013auto}, generative adversarial networks (GANs) \cite{goodfellow2014generative}, and normalizing flows \cite{rezende2015variational} has resulted in investigations on their use for multi-object modeling. The benefits of generative models with respect to multi-object tracking are that they can be used for trajectory prediction \cite{kosiorek2018sequential,jiang2019scalable} or as scene priors for robust handling of occlusion \cite{fernando2018tracking}. Sequential Attend, Infer, Repeat (SQAIR) \cite{kosiorek2018sequential} and a recent follow-up work, SCALOR \cite{jiang2019scalable}, maintain sets of latent variables corresponding to objects in the scene. The latent space of these generative models are structured to make it straightforward to differentiate through the rendering algorithm, allowing for them to be trained to maximize the evidence lower-bound (ELBO) over a dataset of video sequences. Data association is addressed by a ``glimpse'' attention mechanism which sequentially attends to each object in an given frame. Notably, these models can handle objects that enter and leave the scene in the middle of a video sequence and have been applied to multi-pedestrian tracking. Relational-Neural Expectation Maximization \cite{van2018relational} uses iterative inference to assign pixels to object clusters in each image of sequence, and captures interactions between objects using a neural relational dynamics component. 
The iterative inference is necessary to break the symmetry between the latent object components.
R-NEM learns to group the pixels belonging to a particular object to the same latent object component over time, forming a set of object tracks. While these methods are theoretically interesting, an open problem is scaling them to real-world datasets.
Deep generative models have been partially incorporated into existing multi-object tracking frameworks as well. A sequential GAN is used to improve the robustness of a pedestrian tracker in crowded scenes to occlusion and false detections \cite{fernando2018tracking}. They directly generate pedestrian heatmaps with the GAN's generator, which are used to associate new object detections. Then, they maintain a set of tracks by training LSTMs with attention to do short- and long-term trajectory prediction. They demonstrate slightly improved pedestrian detection performance compared to strong baselines on sequences from the PETS2009 benchmark.

To conclude, in this section we reviewed a wide variety of machine learning approaches to the combinatorial optimization aspect of data association. We organized our presentation by describing how each fits into the framework of MDAPs. We first presented search algorithms and non-probabilistic discrete optimization methods to provide context for work done before recent data-driven approaches. Then, we discussed algorithms that fall broadly under the categories of network flow over probabilistic graphs, conditional random fields, belief propagation over factor graphs, MCMC, and end-to-end learning. The end-to-end learning approaches can be contrasted with the other approaches for their abandonment of the structure provided by the combinatorial optimization framework in lieu of an almost complete reliance on data-driven techniques. In the next section, our focus shifts to reviewing recent work whose primary aim is to learn discriminative features for data association that can are used in tandem with some of the algorithms presented in this section.

\section{Learning Features for Data Association}
\label{sec: learning_asgn_costs}
\subsection{Assignment Costs}
The particular choice of the data association cost function can have a large impact on the performance of a downstream task. We can observe from Equations \ref{eq:linear-cost}, \ref{eq:SSMDAP}, and \ref{eq:MSMDAP} that the cost functions for data association measure how ``expensive'' it is to include a particular assignment of detections (or tracks) to tracks in the solution. In this section, we introduce two perspectives towards formulating cost functions, specifically highlighting probabilistic approaches. Following that, we review machine learning methods for learning good features for data association, organized by non-deep learning and deep learning approaches.

\subsubsection{Kinematic Costs}
In situations where sensor measurements only consist of noisy estimates of kinematic data from targets (e.g., position and speed), a probabilistic framework can be used to recover the unobservable state of the targets. The most common approach is to handle the uncertainty in the sensor measurements and target kinematics with a stochastic Bayesian recursive filter; see \cite{mahler2007statistical} for a comprehensive overview. The Kalman Filter--probably the most popular filter of this flavor--provides the means for updating a posterior distribution over the target state given the corresponding measurement likelihood, i.e., $P(x_k \mid z_k) \propto P(z_k \mid x_{k-1}) P(x_{k-1} \mid z_{k-1})$. We are using the same notation as before, such that $x_k$ represents the target state at time $k$ and $z_k$ is the measurement at time $k$. One of the reasons for the popularity of the Kalman Filter is that by assuming that all distributions of interest are Gaussian, the posterior update can be computed in closed form. Recall that a partial association hypothesis $\gamma^j$ for the multi-scan single-sensor data association problem assigns $T$ measurements to a single track within the sliding window of length $T$. The simplest cost function for data association is to minimize the following negative log-likelihood ratio:
\begin{equation}
\label{eq:NLL}
c_{i_1, i_2, ..., i_T} = -\log \frac{ P(\gamma^j \mid z_1^{i}, z_2^{i}, ..., z_T^{i})}{P(\gamma^0 \mid z_1^{i}, z_2^{i}, ..., z_T^{i}) }, \hspace{1em} (\gamma^j, \gamma^0) \in \gamma.
\end{equation}
The partial hypothesis $\gamma^j$ represents the j\textsuperscript{th} track of the hypothesis $\gamma$, and $\gamma^0$ represents a dummy track where all measurements attributed to it are considered false alarms. Assuming the sensor has a probability of 1 of detecting each target and a uniform prior over all assignment hypotheses, the likelihood that the j\textsuperscript{th} track generated the assigned measurements is
\begin{equation}
P(\gamma^j \mid z_1^{i}, z_2^{i}, ..., z_T^{i}) \propto P( z_1^{i}, z_2^{i}, ..., z_T^{i} \mid \gamma^j).
\end{equation}
Assuming independence of the measurements and track states between time steps, we can decompose the likelihood that the measurements originated from track $\gamma^j$ as
\begin{equation}
\label{measurement likelihood}
P( z_1^{i}, z_2^{i}, ..., z_T^{i} \mid \gamma^j) = \prod_{k=1}^{T} P(z_k^{i} \mid x_k) P(x_k \mid j).
\end{equation}
In the Kalman Filter and its extensions, the right-hand side has an attractive closed form representation as a Mahalanobis distance between the measurement predictions and the observed measurements, scaled in each dimension of the measurement space by the state and measurement covariances. This can easily be derived by taking Equation~\ref{measurement likelihood} and plugging it into the negative log-likelihood ratio in Equation~\ref{eq:NLL}.
\par
In track-to-track association, the conventional cost function associated with a partial hypothesis is the likelihood that the tracks from multiple sensors were all generated by the same "true" target. When $S = 2$, the simplest approach is to consider the random variable $\triangle_{12} = x^1 - x^2$, which is the difference between the track state estimates from sensor 1 and sensor 2. When the track state estimates are Gaussian random variables, $\triangle_{12}$ is also Gaussian. The cost function becomes the likelihood that $\triangle_{12}$ has zero mean and covariance given by $\Sigma = \Sigma_1 + \Sigma_2 - \Sigma_{12} - \Sigma_{21}$ \cite{bar2004multisensor}. The first two terms of the covariance are the uncertainty around the track state estimates, and the second two terms are the cross covariances. A straightforward way to extend to the $S \geq 3$ case is to use star-shaped costs $\triangle_{1S} = \sum_{i=2}^{S} \triangle_{1i}$ \cite{walteros2014integer}. For the Gaussian case, the cost can also be written in closed form as a Mahalanobis distance between the track state estimates \cite{kaplan2008assignment} \cite{deb1997generalized}
\begin{equation}
\label{sum-pairwise}
c_{i_1, i_2, ..., i_S} = \sum_{j=2}^{S} \triangle_{1j}^{\intercal}\Sigma_{1j}^{-1}\triangle_{1j}  
\end{equation}
In the Bayesian setting, minimizing Equations \ref{eq:NLL} and \ref{sum-pairwise} is analogous to finding the MAP assignment hypothesis.

\subsubsection{Feature-augmented Costs}
It is often the case in multi-object tracking that sensors generate high-dimensional observations of the surveillance region from which target information must be extracted. The most obvious example of this is the image data generated by a video surveillance system. This data, when featurized, can be used to augment or replace the kinematic costs mentioned in the previous subsection. The goal of doing this is to improve the association accuracy, and ultimately the overall tracking performance.
\par
Due to the high-dimensionality of the raw measurements, almost all such methods attempt to \textit{learn} a pairwise cost between measurements or tracks using features extracted from the data. This pairwise cost can represent the association probability of the two objects, or simply some notion of similarity, e.g., a distance. There are many ways of formulating the problem of learning assignment costs and using it for solving data association or track-to-track association as a machine learning problem. For example, one technique is to use metric learning to transform the high-dimensional sensor measurements into a lower-dimensional geometric space where a Euclidean distance can be used as the assignment cost function. Learning pairwise costs from data is heavily used in the multi-object tracking computer vision community, partially due to the ease at which features can be extracted from images \cite{li2013survey}. Of course, the main question is deciding what features to use, or whether to try to learn the best features for data association directly from data.
\par
There are multiple ways to incorporate learned pairwise costs into data association when viewed as a MDAP. One common approach is as follows. The probability of association for a pair of measurements $\Lambda_i$ and $\Lambda_j$ (or tracks) can be written as a joint pdf \cite{osbome2011track}; assuming independence of the kinematic (K) and non-kinematic (NK) components of this probabilistic cost function, the resulting negative log-likelihood pairwise cost is
\begin{equation}
\begin{aligned}
\label{eq:feature-aug}
  c_{ij} &= -\log P(\Lambda_i , \Lambda_j) \\
 &= -\log \big( P_{\text{K}}(\Lambda_i , \Lambda_j) P_{\text{NK}}(\Lambda_i , \Lambda_j) \big) \\
 &= -\log P_{\text{K}}(\Lambda_i , \Lambda_j) - \log P_{\text{NK}}(\Lambda_i , \Lambda_j).
\end{aligned}
\end{equation}
Usually, $P_{\text{NK}}(\Lambda_i , \Lambda_j)$ is parameterized by weights $\theta$ and is a function of the features extracted from the sensor data and $\theta$. For example, this probability could be represented as a neural network that outputs a similarity score between 0 and 1. The kinematic component of this pairwise cost, $P_{\text{K}}(\Lambda_i , \Lambda_j)$, could be adapted from Equation~\ref{eq:NLL}.

Framing the problem of learning an assignment cost function for data association or track-to-track association is deeply intertwined with the choice of sensor(s). This section will mainly consist of recent work on this problem from the computer vision community, where machine learning is most heavily used. One reason for this is the relatively large amount of annotated video tracking datasets that are available. We divide the presentation of techniques into pre- and post-deep learning to provide a comprehensive perspective and to emphasize the shift to deep learning-based approaches in recent years.

\subsection{Learning Features for Data Association, Pre-Deep Learning}
The goal of learning features for data association is to use (usually labeled) training data to teach a model to output association scores at test time. These scores are then used to compute the assignment costs, as in Equation~\ref{eq:feature-aug}, and these costs are utilized by the optimization frameworks introduced in Section~\ref{sec:opt}. In visual tracking, discriminative models have been commonly trained for predicting association scores based on appearance information. These models are typically adapted from popular classification and ranking models. Another learning paradigm (occasionally used in conjunction with discriminative models) is metric learning. In this case, the goal is to learn a distance metric between measurements or tracks, typically in the form of a parameterized Mahalanobis distance. The next two subsections review these two learning techniques in the context of data association prior to the use of deep learning for feature extraction. As a key challenge for these methods was feature selection, we provide Table~\ref{taxonomy of features} which summarizes the various visual features used for learning association costs. 

\subsubsection{Discriminative models}
Boosting is one of the most powerful techniques in supervised learning and is a natural choice for learning discriminative models that approximate the true association costs. The general idea behind boosting is to produce a series of \textit{weak} learners that are combined to form a single \textit{strong} learner. The HybridBoost algorithm \cite{li2009learning}, one of the first applications of data-driven learning in multi-object tracking, is used to learn the link costs for a network flow graph (Equation~\ref{eq:link-costs}). The data association problem is decomposed into a hierarchy of association problems where the tracklet lengths successively increases \cite{huang2008robust}; furthermore, it is cast as a joint ranking and classification problem. The cost function is learned so that it can rank correct associations higher than incorrect ones, as well as reject some associations entirely (i.e., a binary classification to determine reasonable associations). Hence, HybridBoost is a combination of RankBoost and AdaBoost \cite{freund1995desicion}. Their HybridBoost model is trained offline with videos paired with ground-truth trajectories. In Kuo et al., \cite{kuo2010multi}, a slightly different approach is taken; a hierarchical decomposition is used, but each stage of the hierarchy is linked by applying the Hungarian algorithm and the cost matrix for the Hungarian algorithm is learned online with AdaBoost. Online learning of the discriminative model within the sliding-window is an attractive notion, since variations in appearance at test time can cause difficulty for systems that are trained offline. However, this comes at the cost of potentially sacrificing real-time capabilities. On a task involving tracking 2-8 pedestrians at a time, this tracker runs at about 4 FPS. Other appearance models based on boosting have been proposed where the RankBoost algorithm is used with CRFs \cite{yang2011learning,yang2012online}. In a follow-up work to Kuo et al., ideas from person re-identification are used to improve the appearance model \cite{kuo2011does}. The features used by the boosting algorithms mentioned here are summarized in Table~\ref{taxonomy of features}.
\par
In efforts to improve upon boosting for online learning of appearance models, incremental linear discriminant analysis (ILDA) has been used \cite{bae2014robust}. They showed that ILDA outperforms boosting in their experiments in terms of identification accuracy and computational efficiency, partially due to the fact that ILDA simply requires updating a single LDA projection matrix for distinguishing amongst the appearances of multiple objects. However, this approach makes the assumption that the featurized appearances of the tracked objects can be projected into a vector space where they are linearly separable. The assignment cost they used was 
\begin{equation}
\label{eq:ASM}
c_{ij} = \Lambda(x_i, x_j) = \textcolor{red}{\Lambda^{A}(x_i, x_j)} \textcolor{blue}{\Lambda^{S}(x_i, x_j)} \textcolor{violet}{\Lambda^M(x_i, x_j)}
\end{equation}
for \textcolor{red}{appearance}, \textcolor{blue}{shape}, and \textcolor{violet}{motion} (kinematics) affinities. This form of the cost is similar to Equation~\ref{eq:feature-aug} and is fairly common. The appearance affinity is the score computed by ILDA, and the shape and motion affinities are not learned from data. In this work, tracks are incrementally stitched together from tracklets by repeated application of the Hungarian algorithm. Another alternative to boosting that was explored for learning association costs within complex graphical models was the structured SVM \cite{Kim2013,wang2015learning,wang2017learning,choi2015near}. In general, however, the structured SVM approaches  were restricted to linear cost functions. 

\subsubsection{Metric Learning}
\label{sec:metric-learning}

A different approach to addressing the problems of variability in object appearance is target-specific metric learning. Here, we define metric learning as the problem of learning a distance $d_\mathbf{A}(x,y) = \sqrt{(x - y)^{\intercal}\mathbf{A}(x - y) }$ parameterized by a positive semi-definite (PSD) matrix $\mathbf{A}$. An intuitive way of thinking about this is that the data points $x$, which might be featurized representations of tracked objects, are being mapped to $A^{1/2}x$ where a Euclidean distance metric can be applied to the rescaled data \cite{xing2003distance}. This is then cast as a constrained optimization problem to ensure that the solution $\mathbf{A}$ is valid, i.e., $\mathbf{A} \succeq 0$. An early attempt at applying metric learning in multi-object tracking \cite{wang2010discriminative} combined the problem of learning a discriminative model for appearance matching given image patches with motion estimation and jointly optimized with gradient descent. Their formulation requires running the optimization at each time step for all pairs of objects in the scene with a set of training samples that gets incrementally updated. A more efficient use of metric learning for multi-object tracking is learning link costs in a network flow graph \cite{wang2014tracklet,wang2017tracklet}. A regularized version of the  constrained optimization problem is applied to learn a distance between feature vectors for an appearance affinity model. The intention is to learn a metric that returns a smaller distance for feature vectors within the same tracklet in the graph than for feature vectors that belong to different tracklets. The negative log-likelihood assignment cost for the network links is defined similarly to Equation~\ref{eq:ASM}.

\begin{table}[t]
  \caption{Features used for data-driven learning of assignment costs from a representative set of work.}
  \label{taxonomy of features}
  \centering
  \begin{tabular}{l|l|p{7.5cm}}
    \toprule
    \textbf{Related Work} & \textbf{Method} & \textbf{Summary of Features Used} \\
    \midrule
    \cite{li2009learning} &  HybridBoost & Tracklet lengths, no. of detections in the tracklets, color histograms, frame gap between tracklets, no. of frames occluded, no. of missed detected frames, entry and exit proximity, motion smoothness\\
    \hline
    \cite{kuo2010multi,kuo2011does,yang2012online} & AdaBoost & Color histograms, covariance matrices, HOG  \\
    \hline
    \cite{yang2011learning} & RankBoost & Tracklet lengths, no. of detections in the tracklets, color histograms, frame gap between tracklets, no. of frames occluded, no. of missed detected frames, entry and exit proximity, motion smoothness \\
    \hline
    \cite{bae2014robust} & ILDA & Templates from HSV color channel and tracklet ID \\
    \hline
    \cite{wang2015learning,wang2017learning} & Structured SVM & Off-the-shelf detector confidence (e.g., from DPM \cite{felzenszwalb2010object}), consecutive bounding box IOU, geometric relationships between all pairs of objects \\ 
    \hline
    \cite{wang2014tracklet,wang2017tracklet} & Metric learning & RGB, YCbCr, and HSV color histograms, HOG, two texture features extracted with Schmid and Gabor filters\\
    \bottomrule
  \end{tabular}
\end{table}

\subsection{Learning Features for Data Association, Post-Deep Learning}
Tracking-by-detection is the current state-of-the-art approach for visual tracking, mainly due to the use of CNNs. The basic idea is to first leverage powerful deep networks for object detection to extract raw observations followed by an association step to produce object tracks. In this section, we will discuss the use of CNNs within the data association step.
\par
CNNs learn features directly from raw images which are translation invariant and invariant to slight deformations, removing the need to hand-pick features which may not generalize well. Another reason why deep learning is an attractive option for multi-object tracking is because it is straightforward to take a CNN that has been pretrained on a massive image classification dataset and \textit{transfer} the learned features to new tasks, including estimating association costs. 
\par
One of the first uses of deep learning in multi-object tracking was running image patches of detected objects obtained with, e.g., the DPM \cite{felzenszwalb2010object}, through a CNN to extract features. The CNNs were pretrained on the ImageNet and PASCAL visual object classification (VOC) datasets. In one instance, the features extracted from the CNN were used to train a multi-output regularized least-squares classifier \cite{kim2015multiple}. Essentially, a 4,096-dimensional feature vector is first extracted from a CNN for each detection box, followed by an application of PCA to reduce the dimensionality to 256. The classifier is used to compute a log-likelihood cost for a track hypothesis given a set of detections. This paper was unique in that it showed how the classic multiple hypothesis tracking (MHT) algorithm, which performs MAP inference by updating sets of track hypothesis trees in real-time, compares favorably with the modern approaches described in Section~\ref{sec:opt} when augmented with learned assignment costs. In fact, at the time of publishing, their method (referred to as MHT\_DAM) outperformed the second-best tracker on the 2DMOT15 by 7\% in multiple object tracking accuracy (MOTA). 
\par 
\begin{figure}
\includegraphics[scale=0.18]{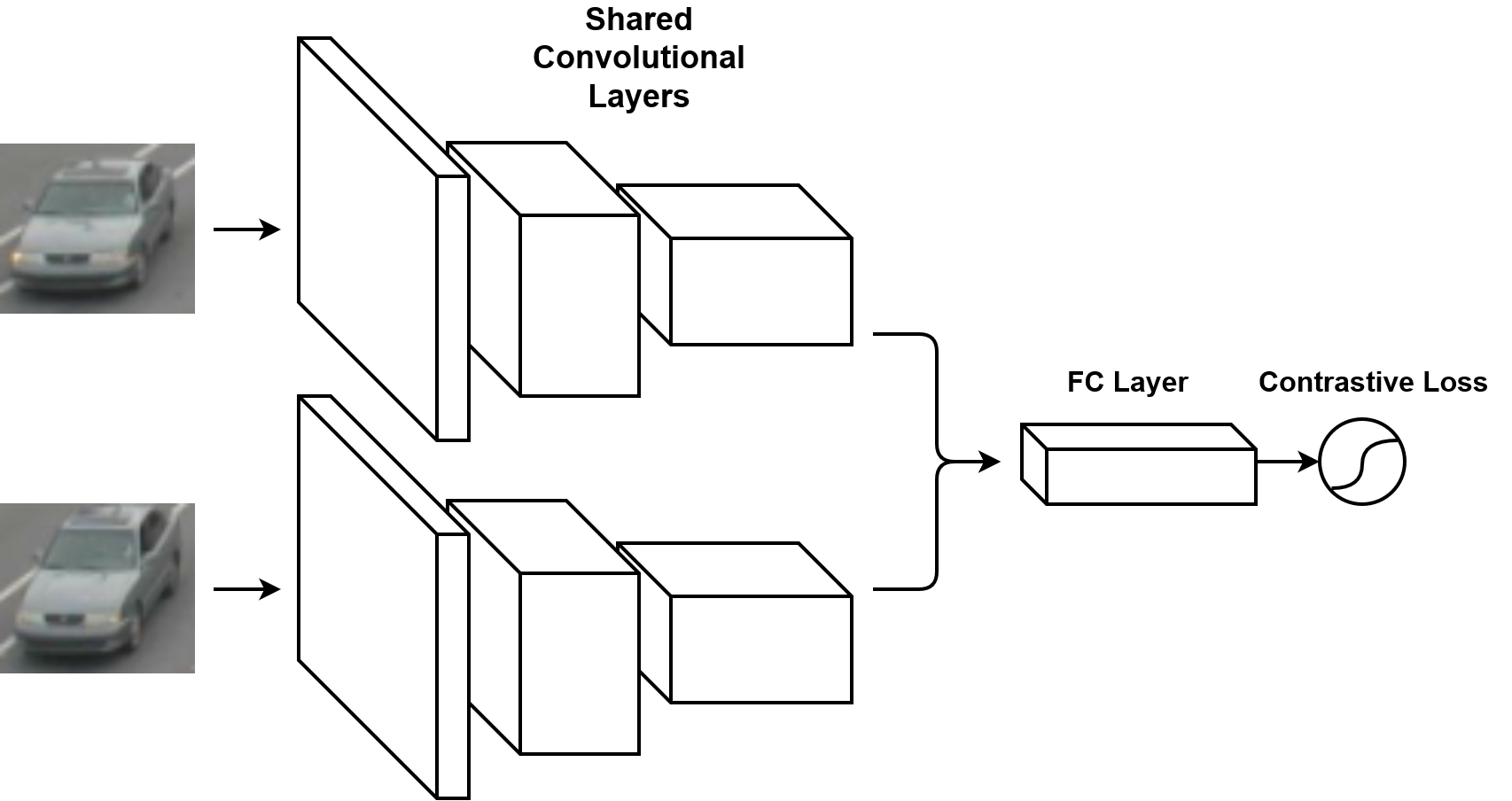}
\caption{The basic architecture of a Siamese network. The weights of the convolutional layers are shared between the two arms of the network. A contrastive loss can be used to train the network to predict the similarity of the two input images.\label{siamese}}
\end{figure}
\subsubsection{Siamese Networks}
A variation on the standard CNN architecture that has seen extensive use in multi-object tracking is the Siamese network. A Siamese network processes two inputs simultaneously using multiple layers with shared weights (Figure~\ref{siamese})  \cite{leal2016learning}. These networks can be used for a variety of tasks that involve comparing two image patches; this seems intuitively useful for the task of learning assignment costs, where we are interested in predicting the association likelihood for two inputs. Indeed, a technique was proposed to directly compute association scores for pairs of image patches  \cite{leal2016learning}. First, two image patches are stacked, along with their optical flow information, and fed as input into a Siamese network. A separate network learns contextual features which encode relative geometry and position variations between the two inputs, and the final layers of these two networks are extracted and combined with a gradient-boosting classifier to produce a match prediction score. Tracks are ultimately obtained by solving a network flow problem (Section~\ref{section:graph-cut}) using linear programming. 
\par
Siamese networks have also been used to learn embeddings for pairs of detections \cite{wang2016joint}. In this work, all parameters between the two arms of the CNN are shared and the features produced by the last layer are used as input to a metric learning loss. Specifically, a multi-task loss function for incorporating temporal constraints is combined with the regularized metric learning loss to jointly optimize the weights of the deep model. They use an online learning algorithm to address the issue of changing object appearance throughout a trajectory, but the deep networks are pretrained with auxiliary data. The learned affinity model is combined with the softassign algorithm \cite{gold1996softmax} to find an optimal pairing of tracklets. For the task of underwater multi-object tracking, Siamese networks were shown to improve performance as well \cite{rahul2017siamese}. Instead of only considering pairs of images with Siamese networks, the Quad-CNN \cite{son2017multi} aims to learn more sophisticated representations for metric learning by considering quadruplets of images. A bounding box regression loss and a multi-task ranking loss that considers appearance and temporal similarities between four images are used to jointly optimize a Quad-CNN end-to-end. The authors propose a sliding window minimax label propagation algorithm for data association. 
\subsubsection{Online Appearance Adaptation}
The confidence-based robust online tracking approach \cite{bae2014robust} has been extended with a deep appearance model \cite{bae2017confidence} resembling a Siamese network. The features from the last CNN layer are used to compute a metric over pairs of image patches such that the metric represents a regularized energy function where the lowest possible energy gets assigned to the optimal assignment hypothesis. They employ online transfer learning  to update a small number of the higher layers in the network to adapt to changing object appearances. When the average affinity scores computed by the network fall below a threshold at runtime, training samples are collected and a pass of online transfer learning is carried out to adapt the network. To help reduce the run-time overhead introduced by online learning, the authors suggest using a parallelized implementation and performing the high-confidence and low-confidence tracklet associations once every 10 time steps, as opposed to every time step. Another efficient online algorithm for updating appearance models has been proposed where a bilinear similarity function is learned between two feature vectors with constrained convex optimization \cite{yang2017hybrid}. The feature vectors are also aggregated from the last layer of a CNN. Ideas from single object tracking and reinforcement learning have been adapted for online multi-object tracking \cite{chu2019online}, where a policy is learned to decide whether the target-specific tracking models should be updated with the latest detections and features at predicted locations provided obtained by ROI pooling.

\subsubsection{Deep Network Flow}
The network flow approach popularized by Zhang et al. \cite{zhang2008global} is revisited again from a deep learning perspective \cite{schulter2017deep,shen2018tracklet}. Effectively, the parameters of the unary and pairwise link costs are learned end-to-end with a deep neural network. The original linear program is converted into the following bi-level optimization problem 
\begin{equation}
\begin{aligned}
& \argmin_{\Theta} \mathcal{L}(x^{gt}, x^*) \\
& \textrm{   s.t.  } x^* = \argmin_{x} c(f, \Theta)^{\intercal} x \\
& \mathbf{A}x \leq \mathbf{b}, \mathbf{C}x = 0
\end{aligned}
\end{equation} 
for parameters $\Theta$, input data $f$, and ground truth network flow solutions $x^{gt}$. The $M$ concatenated flow variables are $x \in \mathbb{R}^M$, and $\mathbf{A} = [\mathbf{I}, -\mathbf{I}]^{\intercal} \in \mathbb{R}^{2M \times M}$ and $\mathbf{b} = [0, 1]^{\intercal} \in \mathbb{R}^M$ are box constraints, and $\mathbf{C} \in \mathbb{R}^{2K \times M}$ are the flow conservation constraints. The inner optimization problem is smoothed so that it is easily solvable with an off-the-shelf convex solver. The high level optimization problem is then solved with gradient descent. The high level optimization problem needs ground truth network flow labels $x^{gt}$ during training; this is handled by manually annotating bounding boxes in sequences of frames. At test time, inference is performed in a sliding window. 

\subsubsection{Other approaches}

A variant of the data association problem for multi-object tracking as a minimum-cost graph multi-cut problem \cite{tang2016multi} has been explored in conjunction with learned features. The key differences here with the previously discussed optimization approaches are that multiple detections at a single time step can be attributed to the same person; also, it is easier to allow edges to connect across multiple time steps in this graph to handle occlusion. The edge costs are learned with logistic regression, with features obtained from the DeepMatching \cite{weinzaepfel2013deepflow} algorithm. DeepMatching uses a CNN that has been trained to produce dense correspondences between image patches, and was notably used in the DeepFlow \cite{weinzaepfel2013deepflow} algorithm for learning large displacement optical flow. It is also used in another multi-object tracking system to compute temporal affinities between input features \cite{henschel2017improvements}. Related to this is recent work on examining the interplay between semantic segmentation and multi-object tracking \cite{milan2015joint,tian2016duality,bullinger2017instance}. 
In particular, a CNN is used to segment images, and then the optical flow between segmented object pairs in consecutive images is used to define an association cost matrix \cite{bullinger2017instance}.

A noticeable trend is a gradual drift away from developing novel optimization algorithms that solve a MDAP; rather, recent methods are relying more on powerful discriminative techniques, such as using features from pretrained CNNs, and combining this with linear assignment solvers. Advances in object detection such as Faster R-CNN \cite{ren2015faster} have almost single-handedly improved the performance of multi-object trackers \cite{ciaparrone2019deep}. A recent paper \cite{bergmann2019tracking} examines this trend in detail and introduces what they refer to as a new multi-object tracking paradigm. They propose to leverage bounding box regression to handle data association with a powerful object detection CNN, and to use a Feature Pyramid Network \cite{lin2017feature} to robustly handle objects of variable sizes. They suggest that it is worthwhile to explore the limits of object detection within multi-object tracking. Their Tracktor model achieves equivalent or convincingly stronger or performance than many state-of-the-art online trackers that use the sophisticated data association methods described in Section~\ref{sec:opt}. 
\par
\begin{figure}
\begin{center}$
\begin{array}{cccc}
\includegraphics[scale=0.33]{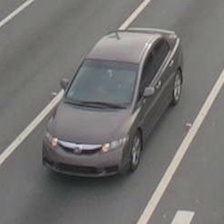} &
\includegraphics[scale=0.33]{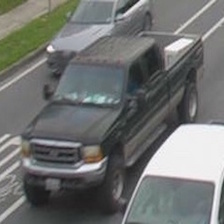} & 
\includegraphics[scale=0.33]{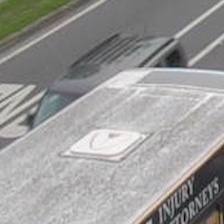} &
\includegraphics[scale=0.33]{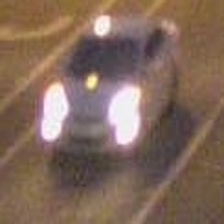} \\
\includegraphics[scale=0.33]{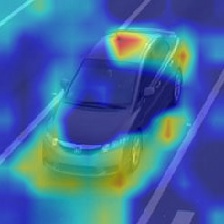} & \includegraphics[scale=0.33]{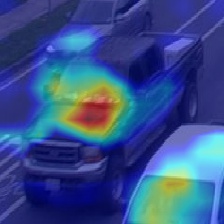} & 
\includegraphics[scale=0.33]{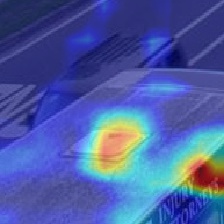} &
\includegraphics[scale=0.33]{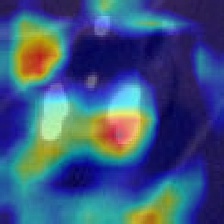} \\
\includegraphics[scale=0.33]{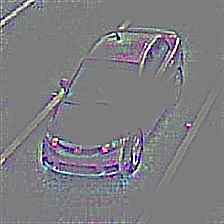} & \includegraphics[scale=0.33]{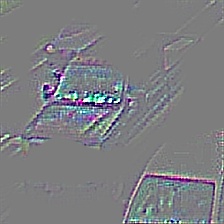} & 
\includegraphics[scale=0.33]{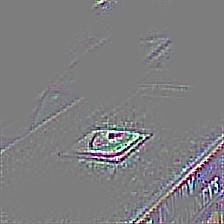} &
\includegraphics[scale=0.33]{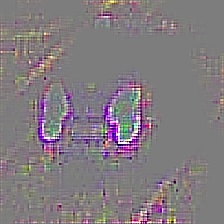}
\end{array}$
\end{center}
\caption{Visualizations of "important" regions for making predictions with the VGG16 network, generated with Grad-CAM \cite{Selvaraju_2017_ICCV} and pre-trained VGG16 weights \cite{Simonyan14c}. The first two images in the top left were correctly labeled as containing vehicles, and it can be seen that the CNN leverages interpretable features such as the car body, tires, and windshield to come to this conclusion. The CNN was not able to correctly classify the vehicles in the two images on the top right. Heavy occlusion and illumination changes still confuse a CNN if it hasn't been trained for these situations. The images were taken with a traffic camera by the authors. Best viewed in color. \label{dl-viz}}
\vspace{-0.1in}
\end{figure}

We would like to provide further insight into the use of CNNs pretrained on image classification datasets for generating detections and learning assignment costs. To this end, we visualized CNN layer activations using the Gradient-weighted Class Activation Mapping technique \cite{Selvaraju_2017_ICCV} in Figure~\ref{dl-viz} when asked to classify images of vehicles at a traffic intersection.

To summarize, in this section we first explained how probabilistic cost functions for data association are formulated with kinematic and non-kinematic components. Then, we reviewed machine learning algorithms for learning association features such as boosting and metric learning. Finally, we discussed a variety of deep learning methods that mainly finetune pretrained CNNs to predict similarity or directly compute association scores. 

\section{Empirical Comparison}
\label{sec: compare}
We have presented a large number of machine learning techniques for data association in multi-object tracking without yet addressing the question of when specific methods may be more preferable than others. We briefly touch on that topic here, focusing on reported results on the 2DMOT15 and MOT17 benchmarks. For an in-depth empirical comparison of deep learning-based multi-object trackers, we direct readers to a survey on this topic \cite{ciaparrone2019deep} and results from the recent 2018 UA-DETRAC competition \cite{lyu2018ua,DETRAC:CoRR:WenDCLCQLYL15}.  For reference, we have provided the results from the MOT15 and MOT17 leaderboards for methods discussed in this survey, organized by data association method, in Tables \ref{tab:benchmark_mot15} and \ref{tab:benchmark_mot17}.











\begin{table*}[t]
\caption{MOT15 challenge results. The symbols $\uparrow$ and $\downarrow$ respectively indicate that higher and lower values are preferred. LA: Linear Assignment, E2E-LA: End-to-end learned LA, E2E-MDAP: End-to-end MDAP, NF: Network Flow, MHT: Multiple-Hypothesis Tracking, CRF: Conditional Random Field }
\label{tab:benchmark_mot15}
\tabcolsep=3pt
\centering
\resizebox{\textwidth}{!}{%
\begin{tabular}{lcccccccccccc}
    \hline
    \textbf{Tracker}                 & \textbf{DA} & \textbf{MOTA $\uparrow$} & \textbf{IDF1 $\uparrow$} & {\textbf{MT $\uparrow$}} & \textbf{ML $\downarrow$} & \textbf{FP $\downarrow$} & \textbf{FN $\downarrow$} & \textbf{ID\_Sw $\downarrow$} & \textbf{Frag $\downarrow$} & \textbf{HZ $\uparrow$} \\ \hline
Tracktor++~\cite{bergmann2019tracking} & LA & 44.1 & 46.7 & 18.0 & 26.2 & 6477 & 26577 & 1318 & 1790 & 0.9 \\

CNNTCM~\cite{wang2016joint}    & LA & 29.6                        & 36.8                   & 11.2                     & 44.0                    & 7786                  & 34733                  & 712                     & 943                    & 1.7                    \\
RAR15pub~\cite{fang2017recurrent} & LA & 35.1 & 45.4 & 13.0 & 42.3 & 6771 & 32717 & 381 & 1523 & 5.4 \\

AMIR15~\cite{sadeghian2017tracking} & LA  & 37.6     & 46.0   & 15.8 & 26.8 & 7933 &  29397 & 1026   & 2024     & 1.0      \\

CDA\_DDALpb~\cite{bae2017confidence}           & LA        & 32.8                                       & 38.8                    & 9.7                    & 42.2                    & 4983         & 35690                  & 614                     & 1583                   & 2.3                    \\
RNN\_LSTM~\cite{milan2017online}  & E2E-LA & 19.0 & 17.1 & 5.5  & 45.6 & 11578 & 36706 & 1490 & 2081 & 165.2\\
MDP~\cite{xiang2015learning} & E2E-LA & 30.3  & 44.7 & 13.0 & 38.4 & 9717 & 32422 & 680 & 1500 & 1.1 \\
MHT\_DAM~\cite{kim2015multiple}       & MHT & 32.4 & 45.3 & 16.0 & 43.8 & 9064  & 32060 & 435  & 826  & 0.7  \\
NOMT~\cite{choi2015near} & CRF & 33.7 & 44.6 & 12.2 & 44.0 & 7762 & 32457 & 442 & 823 & 11.5 \\
DCCRF~\cite{zhou2018deep} & CRF & 33.6 & 39.1 & 10.4 & 37.6 & 5917 & 34002 & 866 & 1566 & 0.1 \\
SiameseCNN~\cite{leal2016learning}      & NF & 29.0     & 34.3    & 8.5 & 48.4 & 5160 & 37798  & 639    & 1316     &  52.8 \\

TSMLCDEnew~\cite{wang2017tracklet} & NF & 34.3 & 44.1 & 14.0 & 39.4 & 7869 & 31908 & 618 & 959 & 6.5 \\
HybridDAT~\cite{yang2017hybrid} & NF & 35.0 & 47.7 & 11.4 & 42.2 & 8455 & 31140 & 358 & 1267 & 4.6 \\
LINF1~\cite{fagot2016improving} & MCMC & 24.5  & 34.8 & 5.5 & 64.6 & 5864 & 40207 & 298 & 744 & 7.5\\
\hline    
\end{tabular}}%
\end{table*}
\begin{table*}[t]
\caption{MOT17 challenge results}
\label{tab:benchmark_mot17}
\tabcolsep=3pt
\centering
\resizebox{\textwidth}{!}{%
\begin{tabular}{lccccccccccc}
    \hline
    \textbf{Tracker}                 & \textbf{DA} & \textbf{MOTA $\uparrow$} & \textbf{IDF1 $\uparrow$} & {\textbf{MT $\uparrow$}} & \textbf{ML $\downarrow$} & \textbf{FP $\downarrow$} & \textbf{FN $\downarrow$} & \textbf{ID\_Sw $\downarrow$} & \textbf{Frag $\downarrow$} & \textbf{HZ $\uparrow$} \\ \hline
Tracktor~\cite{bergmann2019tracking} & LA & 53.5  & 52.3 & 19.5 & 36.6 & 12201 & 248047 & 2072 & 4611 & 1.5 \\
DMAN~\cite{zhu2018online} & LA & 48.2 & 55.7 & 19.3 & 38.3 & 26218 & 263608 & 2194 & 5378 & 0.3 \\
DAN~\cite{sun2019deep} & E2E-LA & 52.4  & 49.5 &21.4 & 30.7 & 25423 & 234592 & 8431 & 14797 & 6.3 \\
DeepMOT~\cite{xu2019deepmot} & E2E-LA & 48.1  & 43.0 & 17.6 & 38.6 & 26490 & 262578 & 3696 & 5353 & 4.9\\
MHT\_DAM~\cite{kim2015multiple} & MHT & 50.7  & 47.2 & 20.8 & 36.9 & 22875 & 252889 & 2314 & 2865 & 0.9 \\
FAMNet~\cite{chu2019famnet} & E2D-MDAP & 52.0  & 48.7 & 19.1 & 33.4 & 14138 & 253616 & 3072 & 5318 & 0.0 \\
\hline    
\end{tabular}}%
\end{table*}
If the tracking task has lots of labeled data available, e.g., pedestrian or vehicle tracking, and real-time performance is not required, currently the approach employed by Tracktor \cite{bergmann2019tracking} of relying heavily on supervised object detection objectively performs best. It saves on the development cost incurred by sophisticated algorithms for stitching together tracklets while maintaining or exceeding their performance. Methods that learn to solve a custom linear assignment and are near real-time tend to score highly (those with data association methods classified as ``LA'' and ``E2E-LA'' in the tables), emphasizing the use of deep learning for extracting both appearance and tracklet features.

If there is relatively little or no labeled data for a particular tracking task, there are certain avenues one can take besides conducting an expensive data collection and labeling effort. Firstly, directly using features from pretrained CNNs without minimal finetuning is still quite effective for estimating association scores. Another option is to leverage pretrained CNNs for dense correspondence or segmentation to extract flow or segmentation features as additional cues for data association \cite{tang2016multi}. While there has been progress on end-to-end unsupervised approaches \cite{fernando2018tracking}, e.g., based on ability to reconstruct the scene, they are still only a promising research direction as opposed to being practically useful.

\section{Conclusions}
\label{sec: conclusions}

In this survey, we argue that viewing data association as an assignment problem helps to conceptualize the large variety of data-driven techniques. We categorized many popular methods that address the combinatorial optimization and feature learning aspects of data association. One of the most exciting research directions that was discussed is the development of methods that attempt to learn the optimization algorithm as well as the features from data. The combinatorial nature of data association and the difficulty of learning a robust similarity metric for objects pose strong challenges, but recent work in this direction is promising. Broadly speaking, a common theme highlighted in this survey is the replacement of more and more parts of the typically cumbersome multi-object tracking pipeline with data-driven modules.

\paragraph{Broader Impacts} Careful consideration is required when deploying these systems out in the real-world. We do not yet have a perfect understanding of when data-driven systems fail, although we already know that such systems tend to reflect (potentially problematic) biases of our society stemming from, for example, the training data \cite{danks2017algorithmic}. 
This is especially important to highlight due to the inherent \textit{dual-use} nature of multi-object tracking \cite{brundage2018malicious}; that is, it has the potential to be used by both benevolent and malicious actors. As smart surveillance systems are increasingly deployed in cities, it is important to be transparent about the capabilities and limitations of current and near-future multi-object tracking. On the other hand, there are many beneficial uses-cases and outcomes for multi-object tracking, such as reducing traffic fatalities, monitoring endangered species, and improving real-time sports analysis. 

\begin{acks}

This work is supported by the \grantsponsor{1}{National Science Foundation}{} under grant \grantnum{1}{1446813} and the \grantsponsor{2}{Florida DOT}{} under grant \grantnum{2}{BDV31-977-45}. Any opinions, findings, and conclusions or recommendations expressed in this material are those of the author(s) and do not necessarily reflect the views of the National Science Foundation. P.M. Pardalos' research is supported by the Paul and Heidi Brown preeminent professorship at ISE, University of Florida.

\end{acks}

\bibliographystyle{ACM-Reference-Format}
\bibliography{survey.bib}

\end{document}